\documentclass{article} 
\usepackage{collas2025_conference,times}
\usepackage{easyReview}

\usepackage{amsmath,amsfonts,bm}

\def\eqref#1{equation~\ref{#1}}

\def\1{\bm{1}}

\DeclareMathAlphabet{\mathsfit}{\encodingdefault}{\sfdefault}{m}{sl}
\SetMathAlphabet{\mathsfit}{bold}{\encodingdefault}{\sfdefault}{bx}{n}

\usepackage{hyperref}
\hypersetup{
    colorlinks=true,
    linkcolor=black,
    filecolor=magenta,
    urlcolor=blue,
    citecolor=black,
    pdftitle={Overleaf Example},
    pdfpagemode=FullScreen,
    }

\usepackage[ruled,vlined]{algorithm2e}

\definecolor{commentcolor}{RGB}{110,154,155}   
\definecolor{keyword}{RGB}{227, 11, 92}
\newcommand{\PyComment}[1]{\ttfamily\textcolor{commentcolor}{\# #1}}  
\newcommand{\PyCode}[1]{\ttfamily\textcolor{black}{#1}} 
\newcommand{\PyKW}[1]{\ttfamily\textcolor{keyword}{#1}}

\usepackage{algpseudocode}
\usepackage{graphicx}
\usepackage{multirow}
\usepackage{rotating}
\usepackage{arydshln}
\usepackage{xcolor}

\definecolor{myyellow}{RGB}{240, 228, 66}
\definecolor{myvermilion}{RGB}{213, 94, 0}
\definecolor{mypink}{RGB}{204, 121, 167}
\definecolor{mygreen}{RGB}{0, 158, 115}
\definecolor{myorange}{RGB}{230, 159, 0}
\definecolor{mycyan}{RGB}{86, 180, 233}
\definecolor{myblue}{RGB}{0, 114, 178}
\definecolor{mygray}{RGB}{128, 128, 128}

\newcommand{\yellow}[0]{\textcolor{myyellow}{yellow}}
\newcommand{\vermilion}[0]{\textcolor{myvermilion}{vermilion}}
\newcommand{\pink}[0]{\textcolor{mypink}{pink}}
\newcommand{\green}[0]{\textcolor{mygreen}{green}}
\newcommand{\orange}[0]{\textcolor{myorange}{orange}}
\newcommand{\cyan}[0]{\textcolor{mycyan}{cyan}}
\newcommand{\blue}[0]{\textcolor{myblue}{blue}}
\newcommand{\gray}[0]{\textcolor{mygray}{gray}}

\title{Reinitializing weights vs units for maintaining plasticity in neural networks}

\author{
\vspace{1mm}
J. Fernando Hernandez-Garcia$^{1}$ \thanks{Corresponding author} , 
Shibhansh Dohare$^{1}$, Jun Luo$^{2}$  \& Richard S. Sutton$^{1,3}$  \\
$^{1}$ Department of Computeing Science, University of Alberta, Edmonton, Alberta, Canada \\
\vspace{1mm}
\texttt{\{jfhernan,dohare,rsutton\}@ualberta.ca} \\
$^{2}$ Huawei Technologies Canada, Edmonton, Alberta, Canada \\
\vspace{1mm}
\texttt{\{jun.luo1\}@huawei.com} \\
$^{3}$ Canada CIFAR AI Chair, Alberta Machine Intelligence Institute (AMII) \\
}

\collasfinalcopy 

\begin{document}

\maketitle

\begin{abstract}
Loss of plasticity is a phenomenon in which a neural network loses its ability to learn when trained for an extended time on non-stationary data.
It is a crucial problem to overcome when designing systems that learn continually.
An effective technique for preventing loss of plasticity is reinitializing parts of the network.
In this paper, we compare two different reinitialization schemes: reinitializing units vs reinitializing weights.
We propose a new algorithm, which we name \textit{selective weight reinitialization}, for reinitializing the least useful weights in a network. 
We compare our algorithm to continual backpropagation and ReDo, two previously proposed algorithms that reinitialize units in the network.
Through our experiments in continual supervised learning problems, we identify two settings when reinitializing weights is more effective at maintaining plasticity than reinitializing units: (1) when the network has a small number of units and (2) when the network includes layer normalization.
Conversely, reinitializing weights and units are equally effective at maintaining plasticity when the network is of sufficient size and does not include layer normalization. 
We found that reinitializing weights maintains plasticity in a wider variety of settings than reinitializing units.
\end{abstract}

Systems that learn from a continuous stream of data, that \textit{learn continually}, are better suited for making predictions about a changing world such as ours. 
For example, a system that learns continually in a water treatment plant makes more accurate predictions than a system that learns offline and is then deployed \citep{janjua2023gvfs}.
Similarly, a system that continually adjusts its predictions about drivers' earnings makes better ride-sharing matches than alternatives based on fixed heuristics \citep{azagirre2024better}.
Even current large language model systems such as ChatGPT \citep{gpt4} could be improved if they are designed to learn continually; such systems could stay updated with current information without needing to be retrained from scratch.
The already impressive performance of modern deep learning systems could be further enhanced if these systems are designed to learn continually.

However, modern deep learning systems were designed using the train-once approach, in which networks are trained once on a large dataset, then frozen and deployed. 
Unfortunately, the techniques developed under the train-once approach are often unsuccessful in continual learning.
A form of failure of conventional deep learning systems is the loss of the ability to learn when the system is trained for an extended time on non-stationary data, a phenomenon known as \textit{loss of plasticity} \citep{dohare2024loss}.
Since the essential requirement of a learning system is that it is capable of learning from data, the loss of plasticity presents a fundamental problem for deep learning systems that continually learn.

Fortunately, loss of plasticity can be prevented.
An effective and straightforward technique for mitigating loss of plasticity is sporadically reinitializing parts of the network. 
Reinitialization algorithms must carefully balance maintaining plasticity and preserving the information stored in the network's weights. 
If a large part of the network is reinitialized at once, the information necessary for making correct predictions may be destroyed, which can harm performance.
On the other hand, if reinitialization is performed too infrequently, the network may still suffer from plasticity loss.
Continual backpropagation \citep{dohare2021continual, dohare2024loss} and ReDo \citep{sokar2023dormant} achieve this balance by occasionally reinitializing the least useful units or dormant units, respectively.

The idea of reinitializing parts of the network can be implemented at many different levels, such as the entire network \citep{nikishin2022primacy}, a number of layers \citep{nikishin2022primacy, dohare2024loss}, units \citep{dohare2021continual, sokar2023dormant, dohare2024loss}, or weights in the network, we refer to each of these as reinitialization schemes. 
Of all the reinitialization schemes, reinitialization at the level of the weights has yet to be studied for the purpose of maintaining plasticity.
This paper fills this gap in the literature by proposing a new algorithm for reinitializing weights in the network named \textit{selective weight reinitialization algorithms}.
Every certain number of updates, selective weight reinitialization measures the utility of the weights in each layer of the network and reinitializes a fraction of the weights with the lowest utility.

Using selective weight reinitialization, continual backpropagation, and ReDO, we empirically investigate whether there are settings where reinitializing weights is more effective than reinitializing units for maintaining plasticity.
We first study this question with fully-connected networks in the permuted MNIST problem \citep{goodfellow2013empirical, zenke2017continual}, where we found that reinitializing weights is more effective at maintaining plasticity in two settings: (1) when the network has a small number of units per layer and (2) when the network employs layer normalization. 
We then proceed to compare both reinitialization schemes in a class-incremental learning problem based on the CIFAR-100 dataset using vision transformers \citep{dosovitskiy2021an}.
There, we find that no algorithm is particularly successful unless using a reparameterized version of layer normalization, in which case, selective weight reinitialization is the most successful at maintaining plasticity. However, it does not entirely fix the problem.
Continual backpropagation and ReDo are also effective at maintaining plasticity, but do not always result in stable learning. 
Overall, we found that reinitializing weights successfully maintained plasticity in a wider variety of settings than reinitializing units, suggesting that it is a more reliable reinitialization scheme in continual supervised learning.

Our study reveals settings where the well-studied reinitialization scheme, reinitializing units, loses plasticity. 
We contribute towards a general solution for maintaining plasticity in neural networks by proposing a new reinitialization scheme that reinitializes weights. 
This new scheme prevents loss of plasticity in settings where reinitializing units fails to maintain plasticity.
In addition to maintaining plasticity, reinitializing weights has the added benefit of being straightforward to implement.
Measuring the utility of units in a network architecture must account for the complex connectivity patterns of the different structures within the network.
On the other hand, reinitializing weights does not require accounting for complex interdependencies between structures, making it readily applicable to any network architecture.

\vspace{-10pt}
\section{Related Work}
\label{sec:related}
\vspace{-10pt}
\subsection*{Loss of plasticity}
Recently, the loss of plasticity has drawn the attention of the machine learning community.
At first, the observations were presented in different subfields in machine learning, such as class-incremental learning \citep{chaudhry2018reimannian}, supervised learning \citep{ash2020warm}, reinforcement learning \citep{dohare2021continual, nikishin2022primacy, lyle2022understanding}, and continual learning \citep{dohare2020interplay, rahman2021toward}, but the observations were not recognized as part of the same underlying phenomenon.
However, it was only when \citeauthor{rich2022maintaining} (\citeyear{rich2022maintaining}) presented a direct study of the phenomenon that the community developed a unifying language, and all previous observations were attributed to the same underlying phenomenon.
Since then, an increasing number of papers have studied the loss of plasticity effect in the last couple of years \citep{abbas2023loss, sokar2023dormant, lyle2023understanding, lyle2024normalization, lee2024slow, lee2024plastic, elsayed2024addressing, elsayed2024weight, dohare2024loss, lewandowski2024learning, kumar2024maintaining}.

Along with the direct study of loss of plasticity, several algorithms have been proposed to mitigate the effect.
Proposed techniques for maintaining plasticity include regularizing the parameters of the network \citep{kumar2024maintaining, lewandowski2024learning, dohare2024loss, elsayed2024weight}, architectural modifications \citep{lyle2023understanding, abbas2023loss, nikishin2023deep, lyle2024normalization, lee2024slow}, adding parameter noise \citep{ash2020warm, elsayed2024addressing}, and, the focus of this paper, reinitialization techniques \citep{nikishin2022primacy, dohare2021continual, sokar2023dormant, dohare2024loss}. 
Moreover, combining multiple techniques is often more effective at maintaining plasticity than any of them alone \citep{lee2024plastic, dohare2024loss}.
We contribute to this rich literature by proposing a novel reinitialization algorithm that maintains plasticity across a wide variety of settings. 
Our algorithm can be easily applied to any architecture and combined with other methods for maintaining plasticity. 

In addition to their benefits in continual learning, plasticity-preserving algorithms may also help train current deep learning systems, such as large language models. 
For example, a recent paper by \citeauthor{springer2025overtrained} (\citeyear{springer2025overtrained}) found that extended pre-training of language models can negatively impact fine-tuning performance.
These results are reminiscent of the work done by \citeauthor{ash2020warm} (\citeyear{ash2020warm}), who found that pre-training on a small subset of a dataset can lead to lower performance when later fine-tuning on the full dataset. 
Their algorithm, shrink and perturb, remedied the problem, and it is likely to help remedy the poor fine-tuning performance reported by \citeauthor{verwimp2025same} (\citeyear{verwimp2025same}). 
This possibility becomes more likely given that other work has found that using shrink and perturb doubled the convergence speed during fine-tuning of pre-trained models \citep{verwimp2025same}.

\subsection*{Reinitialization algorithms}

Reinitialization algorithms have been employed for maintaining plasticity \citep{nikishin2022primacy, dohare2021continual, sokar2023dormant, dohare2024loss} and for improving generalization performance in neural networks \citep{mahmood2013, taha2021knowledge, alabdulmohsin2021impact, zhou2022fortuitous, zaidi2023when}.
Reinitializing layers in a network has been shown to increase the decision margins and promote convergence to a flatter local minimum, resulting in improved generalization in supervised learning \citep{alabdulmohsin2021impact}.
For loss of plasticity, reinitialization has been used to restart dormant or dead units in the network and restore the initial conditions of the weights that promote learning \citep{sokar2023dormant, dohare2024loss}.
These algorithms vary in what parts of the network they reinitialize, such as the entire network, groups of layers, single layers, and units.

Reinitialization is also an integral part of dynamic sparse training algorithms. 
While the primary goal of such algorithms is to directly learn a sparse network, the algorithms often involve pruning and restarting weights in the network \citep{mocanu2018scalable, evci2020rigging}. 
The motivation behind dynamic sparse training algorithms is to explore the space of subnetworks within a larger network to find a sparse solution \citep{frankle2018the}.
These algorithms have been shown to be robust to periodic changes in their input distribution, suggesting that they may also be effective for maintaining plasticity \citep{grooten2023automatic}.
The algorithm introduced in this paper, selective weight reinitialization, has parallels with dynamic sparse training algorithms.
However, instead of learning a sparse network, we entirely focus on reinitializing weights to maintain plasticity. 
An interesting avenue for future work is the development of an algorithm that trains a sparse network while maintaining plasticity simultaneously. 

Finally, there is a biological basis for reinitialization algorithms.
Biological neurons have been observed to periodically prune a proportion of their synaptic connections, while simultaneously growing new connections at the same rate \citep{kasai2021spine}.
This process is analogous to the continuous initialization of weights in reinitialization algorithms.
The synaptic pruning and growing process in biological neurons suggests that reinitialization may be a requirement to facilitate continual learning in connectionist networks.
Notably, reinitialization occurs at the level of synaptic connections, equivalent to weights in neural networks, rather than at the level of neurons.

\section{Learning problem}
\label{sec:problem}

We limit our study of plasticity to the continual supervised learning setting. 
In this setting, a learning system generates predictions, $\hat{\boldsymbol{y}} \in \mathbb{R}^c$, based on observations, $\boldsymbol{x} \in \mathbb{R}^n$, to match a target, $\boldsymbol{y} \in \mathbb{R}^c$.
Observations and targets are sampled jointly from a probability distribution $p$, which changes every certain number of samples, $N$.
For convenience, we refer to all the observation-target pairs sampled from the same probability distribution as a task.
We subscript the probability distribution of each task by $k$.
Thus, observation-target pairs are sampled according to $p_0$ in the first task, $p_1$ in the second task, and so on.
On each task, the goal of the learning system is to minimize the expected loss between its predictions and the targets $\mathbb{E}_{p_{k}} \left[ \ell (\boldsymbol{y}, \boldsymbol{\hat{y}}) \right]$. 
For the rest of the paper, we consider supervised classification with targets $\boldsymbol{y}$ as one-hot vectors, and $\hat{\boldsymbol{y}}$ as probability distributions. 
For loss, we use the cross-entropy loss $\ell (\boldsymbol{y}, \boldsymbol{\hat{y}}) = -\sum^c_{j=1} y_j \log ( \hat{y}_j)$.

We use a neural network parameterized by $\boldsymbol{\theta}$ to generate predictions in the continual supervised learning setting, $f_{\boldsymbol{\theta}} (\boldsymbol{x}) = \hat{\boldsymbol{y}}$.
At learning step $t \in \{0, 1, \dots, N - 1\}$ in a task, the network receives a mini-batch of $m$ observation-target pairs, $\{(\boldsymbol{x}_i, \boldsymbol{y}_i) \}^m_{i=1}$, sampled from the probability distribution of the current task, $p_k$ for $k \geq 0$. 
To keep track of the evolution of the learning system, we subscript the parameters of the network by the current learning step and the current task number, $\boldsymbol{\theta}_{N \cdot k + t}$.
Since access to $p_k$ is not often available, the network parameters are updated to minimize the empirical loss $\hat{J}(\boldsymbol{\theta}_{N \cdot k + t}) = \frac{1}{m}\sum^m_{i=1} \ell (\boldsymbol{y}_i, f_{\boldsymbol{\theta}_{N \cdot k + t}}(\boldsymbol{x}_i))$ based on the current mini-batch of data.
To update the network parameters, we use the stochastic gradient descent rule,
\begin{equation*}
    \label{eq:sgd_update}
    \boldsymbol{\theta}_{N \cdot k + t + 1} \overset{.}{=} \boldsymbol{\theta}_{N \cdot k + t} - \alpha \nabla_{\boldsymbol{\theta}_{N \cdot k + t}} \hat{J}(\boldsymbol{\theta}_{N \cdot k + t}),
\end{equation*}
where $\alpha \geq 0$ is a step size parameter that scales the size of the update and $\nabla_{\boldsymbol{\theta}_{N \cdot k + t}} \hat{J}(\boldsymbol{\theta}_{N \cdot k + t})$ is the gradient of the empirical loss with respect to the current network parameters.

To measure the loss of plasticity, we compare the learning performance in the current task of a network trained continually on all previous tasks with that of a newly initialized network.
If the performance of the network trained continually is lower than the performance of the newly initialized network, then the network trained continually has lost plasticity. 
In the permuted MNIST experiments in Sections \ref{sec:swr} and \ref{sec:permuted_mnist}, we use the accuracy computed as the network is learning as a measure of performance, \textit{online accuracy}.
In the incremental CIFAR-100 experiments in Section \ref{sec:lop_in_vit}, we use the accuracy computed on a separate test set, \textit{test accuracy}.
Both of these metrics measure the network's ability to generalize to unseen data, which differs from the ability to minimize the loss studied in other papers \citep{lyle2023understanding, elsayed2024addressing}.  
Henceforth, we use loss of the ability to generalize and loss of plasticity interchangeably. 
Still, we note that loss of plasticity has been used to refer to both loss of trainability \citep{lyle2023understanding, lewandowski2024learning} and loss of generalizability \citep{ash2020warm, lee2024slow, dohare2024loss}.

\section{Reinitializing weights for maintaining plasticity}
\label{sec:swr}

Several reinitialization schemes have been proposed for the purpose of maintaining plasticity.
However, one remains to be explored in the loss of plasticity literature: reinitializing weights.
The first contribution of this paper is to propose an algorithm that reinitializes weights and to study its effectiveness in maintaining plasticity.

We named our algorithm \textit{selective weight reinitialization}.
Every certain number of updates, selective weight reinitialization measures the utility of the weights in the network and reinitializes a fraction of the weights with the lowest utility. 
The motivation for reinitializing parts of the network is to restore the initial conditions that enabled the network to learn, which were gradually removed during the learning process.
The algorithm involves five different design choices: the utility function, $U$, used for ranking the weights; the pruning function $P$, which removes low-utility weights from the network; the reinitialization method, $\mathcal{R}$, which dictates how to reinitialize parameters in the network; the reinitialization frequency, $\tau$; and the reinitialization factor, $k$, which dictates how many weights to prune and reinitialize on each reinitialization step.

\textbf{Utility Functions $U$.} 
We study two utility functions: a utility function based on the magnitude of the weights, \textit{magnitude utility}, and a utility function that utilizes the product of the weight magnitude and the magnitude of the gradient with respect to the weights, which we call \textit{gradient utility}.
Given a weight, $w$, in a matrix, $\boldsymbol{W}$, the magnitude utility function assigns a utility of $\left| w \right|$ to the weight. 
The motivation for this utility is simple. 
Large weights are more likely to significantly impact the network's output; thus, removing weights with low magnitude utility is expected to result in minimal changes in the output.
The gradient utility function assigns a utility of $\left|w \cdot g_w \right|$, where $g_w$ is the derivative of the loss with respect to $w$, which can be estimated from a mini-batch of data.
This second utility function is a first-order Taylor approximation of the absolute change in loss when the value of $w$ is set to zero.
Weights with low gradient utility have the least impact on the loss function, so it is likely safe to remove them without incurring a significant change in loss.
Both utility functions are widely used in neural network pruning, yielding comparable results \citep{blalock2020state}, and can be implemented with minimal computational overhead.

\begin{algorithm}[b!]
\SetAlgoLined
    \PyComment{utilities: vector of utilities} \\
    \PyComment{k: reinitialization factor} \\
    \PyKW{def} \PyCode{proportional\_pruning(utilities, k):} \\
    \Indp  
        \PyCode{indices = argsort(utilities)} \PyComment{sorted indices based on utilities}  \\
        \PyCode{num\_pruned = int(k * len(utiltities)) + bernoulli(k * len(utilities))} \\
        \PyKW{return} \PyCode{indices[:num\_pruned]} \\
    \Indm 
    \PyCode{}\\
    \PyKW{def} \PyCode{threshold\_pruning(utilities, k):} \\
    \Indp  
        \PyCode{threshold = k * average(utilities)}   \\
        \PyComment{find indices where the utility is lower than or equal to the threshold} \\
        \PyCode{indices = where(utilities <= threshold)}   \\
        \PyKW{return} \PyCode{indices} \\
    \Indm 
\caption{Proportional and Threshold Pruning (python-like pseudocode)}
\label{alg:pruning_functions}
\end{algorithm}

\textbf{Pruning Functions $P$.}
We utilize the utility of weights in two different ways to prune and subsequently reinitialize the weights in the network.
In the first pruning function, we remove a proportion of the weights with the lowest utility. 
Given a weight vector $\boldsymbol{w} \in \mathbb{R}^d$ and a reinitialization factor $k \in (0,1)$, we prune and reinitialize $k \cdot d$ weights on each reinitialization step.
To handle the decimal part of the product, we sample from a Bernoulli distribution with probability equal to the decimal part. 
Thus, the number of reinitialized weights is: $\text{Integer Part}\left(k \cdot d) +  \text{Bernoulli}(\text{Decimal Part}(k \cdot d)\right)$.
We call this type of pruning \textit{proportional pruning}.
Aside from how we handle the decimal part, proportional pruning is akin to how dynamic sparse training algorithms prune weights \citep{mocanu2018scalable}.
In the second pruning function, we prune all the weights whose utility is below a certain threshold.
Specifically, for a reinitialization factor $k > 0$, we prune all the weights $w$ with utility lower than $k \cdot \overline{U}(\boldsymbol{w})$, where $\overline{U}(\boldsymbol{w})$ is the sample average of the utility of all the weights in the layer, $\frac{1}{d}\sum^d_{i=1} U(w_i)$.
This function is used in ReDo, which prunes and reinitializes units in a network with an activation threshold below or equal to a specific threshold \citep{sokar2023dormant}. 
We call this type of pruning \textit{threshold pruning}.
The key difference between these two types of pruning is that proportional pruning removes, on average, the same number of weights on each reinitialization step, whereas threshold pruning removes a varying number of weights. 
We give pseudocode for these two functions in Algorithm \ref{alg:pruning_functions}.

\textbf{Reinitialization Methods $\mathcal{R}$.}
We devise two reinitialization methods based on the initialization distribution used at the start of training.
The first reinitialization method samples new values from the initialization distribution.
For example, if the entries of a matrix, $\boldsymbol{W}$, were initialized according to a Normal distribution with mean $\mu$ and standard deviation $\sigma$, then when reinitializing $w \in \boldsymbol{W}$, we sample its new value from $\mathcal{N}(\mu, \sigma)$.
On the other hand, if the entries of the bias vector, $\boldsymbol{b}$, were initialized to a fixed value of zero, then at reinitialization $b \in \boldsymbol{b}$ would be set to zero. 
We refer to this reinitialization method as \textit{resample reinitialization} because new weights are resampled from the initial distribution.
We chose this method because it moves the distribution of weights closer to the initialization distribution, which is designed to facilitate learning \citep{glorot2010understanding, he2015delving}.
The second reinitialization method reinitializes weights to the mean of their initialization distribution.
In the same example as before, $w$ would be reinitialized to $\mu$, and $b$ would be reinitialized to zero.
We refer to this reinitialization method as \textit{mean reinitialization}.
Since the mean of initialization functions is often zero, this method is equivalent to setting the value of new weights to zero.
Setting the values of new connections to zero yields good generalization performance in dynamic sparse training algorithms \citep{mocanu2018scalable, evci2020rigging}.

Note that a reinitialization function would be required for each weight matrix and bias vector in the network for either of these reinitialization methods.
We represent a reinitialization method as a set of reinitialization functions, one for each weight matrix and bias vector.
Thus, for a network parameterized by $\left\{\boldsymbol{W}_1, \boldsymbol{W}_2, \dots \right\}$, omitting bias vectors for simplicity, the corresponding reinitialization method is $\mathcal{R} = \left\{I_1, I_2, \dots \right\}$, where $I_i$ is the reinitialization function for a matrix $\boldsymbol{W}_i$.
Given a utility function, a pruning function, and a reinitialization method, the reinitialization frequency, $\tau$, and reinitialization factor, $k$, are treated as hyper-parameter values to be tuned.
Algorithm \ref{alg:swr} gives the pseudocode for selective weight reinitialization. 
The code used for running all the experiments is available at \url{https://github.com/JFernando4/collas_2025_swr_paper}.

\begin{algorithm}[t]
\SetAlgoLined
    \PyComment{weights: list of neural network weight vectors [W\textsubscript{1}, ..., W\textsubscript{l}]} \\
    \PyComment{U: element-wise utility function} \\
    \PyComment{P:  proportional or threshold pruning function} \\
    \PyComment{R: list of reinitialization functions [I\textsubscript{1}, ..., I\textsubscript{l}]} \\
    \PyComment{$\tau$: reinitialization frequency} \\
    \PyComment{k: reinitialization factor} \\
    \PyKW{for} \PyCode{step} \PyKW{in} \PyCode{range(training\_steps):} \\
    \Indp   
        \PyComment{sample mini-batch}   \\
        \PyComment{compute predictions, loss, gradients, and update network} \\
        \PyKW{if} \PyCode{(step + 1) \% $\tau$ == 0:}\\
        \Indp
            \PyKW{for} \PyCode{i} \PyKW{in} \PyCode{range(len(weights)):} \\
            \Indp
                \PyCode{utilities = U(weights[i])} \hspace{34pt} \PyComment{compute utilities} \\
                \PyCode{indices = P(utilities, k)} \hspace{34pt} \PyComment{indices of pruned weights} \\
                \PyCode{new\_weights = R[i](len(indices))} \PyComment{reinitialize weights}\\
                \PyCode{weights[i][indices] = new\_weights} \\
            \Indm
        \Indm
    \Indm 
\caption{Selective Weight Reinitialization (python-like pseudocode)}
\label{alg:swr}
\end{algorithm}

\textbf{Initial assessment.}
Our first goal is to identify which of the many variants of selective weight reinitialization is effective at maintaining plasticity. 
We use the permuted MNIST problem \citep{goodfellow2013empirical, zenke2017continual}, which consists of several tasks, each corresponding to different random permutations of the pixels of the images of the MNIST dataset.
We train networks on 1,000 different permutations with a mini-batch size of 30.
For each permutation, we only do one pass through the data, resulting in 2,000 updates to the network per task.
We use a fully-connected network with ReLU activations, three hidden layers and 100 units per layer, trained using stochastic gradient descent.

We use three different baselines. 
First, the system without any added components, which we refer to as the \textit{base system}.
Second, the base system trained using \textit{L2-regularization}, which adds a penalty term to the loss function proportional to the squared L2-norm of the parameter vector. 
Lastly, we use a baseline that adds parameter noise to the L2-regularization system on every parameter update, a method proposed by \citeauthor{ash2020warm} (\citeauthor{ash2020warm}) which they called \textit{shrink and perturb}.
\begin{figure}[t!]
    \centering
    \includegraphics[width=\linewidth]{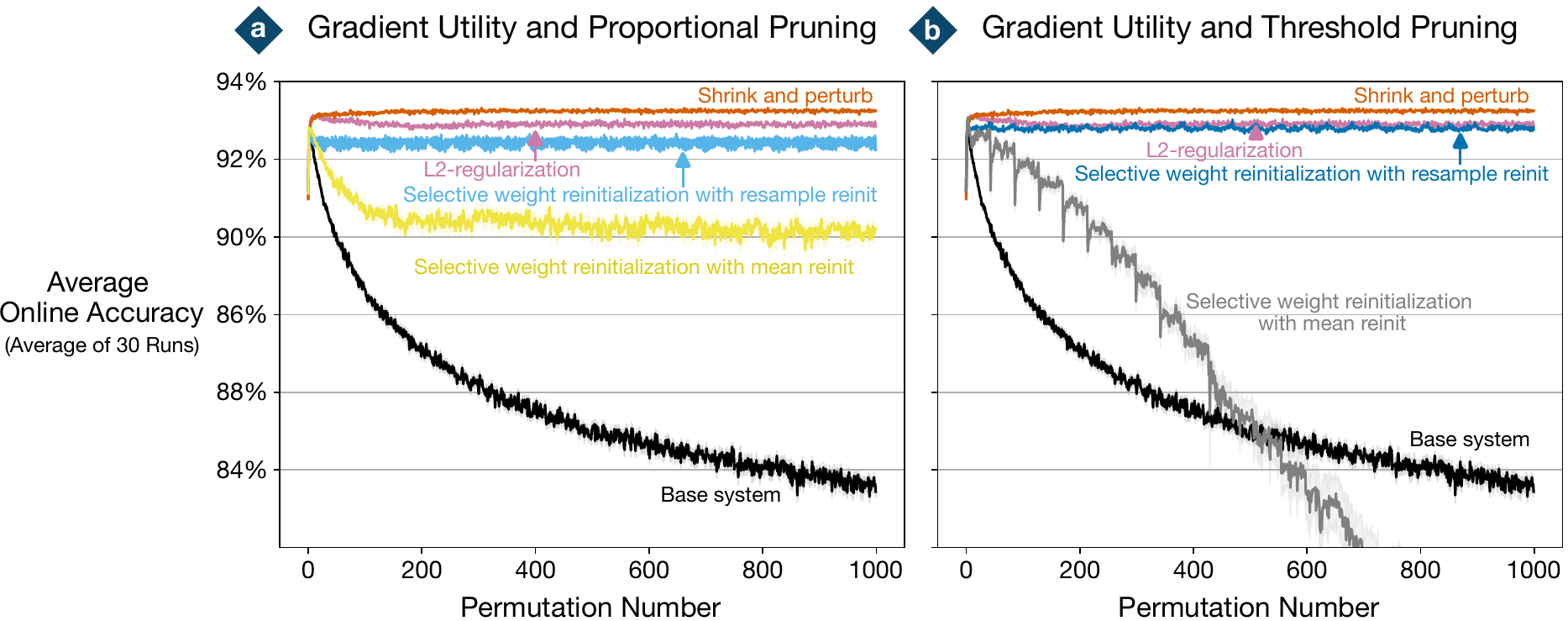}
    \caption{
    Average online accuracy of selective weight reinitialization with gradient utility and (\textbf{a}) proportional or (\textbf{b}) threshold pruning. 
    Each line is the average of 30 runs, while the shaded regions correspond to the standard error.
    All variants of selective weight reinitialization had higher average online accuracy than the base system.
    However, only selective weight reinitialization with resample reinitialization maintained plasticity.
    }
    \vspace{-10pt}
    \label{fig:swr_first_assessment}
\end{figure}
L2-regularization and shrink and perturb have been reported to be strong baselines in the permuted MNIST problem \citep{kumar2024maintaining, dohare2024loss}.
These three baselines serve as a reference for a system that loses plasticity, the base system, and two systems that maintain plasticity: L2-regularization and shrink and perturb. 

As the network learns, we measure its accuracy before it updates from a mini-batch; we refer to this measurement as the \textit{online accuracy}.
As a measure of performance, we use the online accuracy averaged over all the mini-batches of each task---the \textit{average online accuracy}.
Because we only perform one pass through the data per permutation, the average online accuracy serves as a measure of generalization.
Figure \ref{fig:swr_first_assessment} shows the average online accuracy for each task of each algorithm.
A downward trend in performance indicates a loss of plasticity, a flat line indicates that the system maintains plasticity, and an upward trend indicates positive transfer between tasks.  
We only include the results using gradient utility in the main text; see Appendix \ref{app:mnist_extended} for results using magnitude utility. 

The base system (black) loses plasticity over time, resulting in a steady decrease in performance.
On the other hand, L2-regularization (\pink) and shrink and perturb (\vermilion) maintain plasticity; their performances remain mostly flat. 
Selective weight reinitialization yields mixed results, depending on the reinitialization method.
With resample reinitialization (\cyan \ and \blue), selective weight reinitialization maintained plasticity, sometimes matching the performance of L2-regularization.
In contrast, mean reinitialization (\yellow \ and \gray) had a decrease in performance, signalling plasticity loss.  
We present further analysis involving correlates of loss of plasticity in Appendix \ref{app:correlates_mnist}.

\textbf{Takeaways.} 
Gradient utility and resample initialization successfully maintained plasticity with both proportional and threshold pruning. 
Between the two pruning functions, threshold pruning performed slightly better.
Henceforth, we present the results of selective weight reinitialization with gradient utility, threshold pruning, and resample reinitialization, and also provide extended results in the appendices.

\section{Reinitializing weights vs reinitializing units for maintaining plasticity in fully-connected networks}
\label{sec:permuted_mnist}       

We use selective weight reinitialization to study the question: are there settings where reinitializing weights is more effective at maintaining plasticity than reinitializing units?
We use two different algorithms for reinitializing units: continual backpropagation and ReDo.
Continual backpropagation reinitializes low-utility units in the network at a fixed replacement rate.
Newly reinitialized units are protected from being reinitialized again for several updates until reaching a maturity threshold.
On the other hand, ReDo reinitializes units whose activity falls below a certain threshold at a specified reinitialization frequency. 
Continual backpropagation reinitializes units at a fixed rate, analogous to proportional pruning in the previous section. 
In contrast, ReDo reinitializes a varying number of units in each reinitialization step, which is analogous to threshold pruning. 
We employ selective weight reinitialization with gradient utility, threshold pruning, and resample reinitialization to reinitialize weights. 

We devise two settings where reinitializing units may fail to balance maintaining plasticity and preventing the loss of previously learned information.
The first setting is when the network architecture includes layer normalization.
Layer normalization is a technique for normalizing the values of units in a layer by subtracting the sample average and dividing by the sample standard deviation \citep{ba2016layernormalization}.
In previous correspondence with the authors of continual backpropagation \citep{dohare2021continual}, they mentioned that the algorithm did not work well when combined with layer normalization; however, a systematic study of this observation was never conducted.
For the second setting, we use networks with a small number of units.
Intuitively, small networks have less capacity to spare for reinitializing parts of the network; hence, reinitializing a large portion of the network could harm performance. 
Since selective weight reinitialization works at a smaller level of granularity, we hypothesize that it would be a more effective reinitialization scheme than reinitializing units.

\begin{figure}[t!]
    \centering
    \includegraphics[width=\linewidth ]{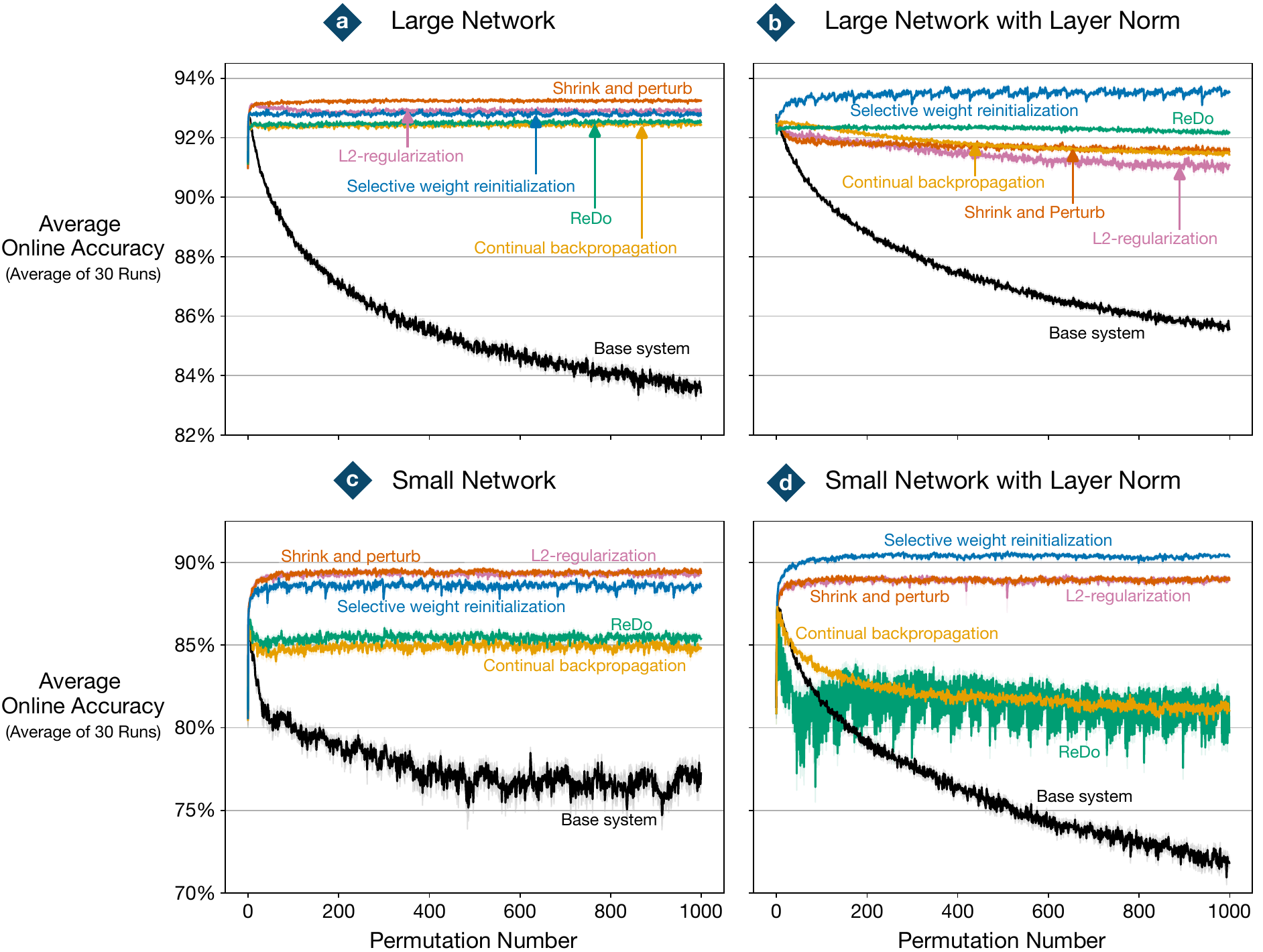}
    \caption{
    Average online accuracy of selective weight reinitialization, ReDo, and continual backpropagation in four settings: (\textbf{a}) large network, (\textbf{b}) large network with layer norm, (\textbf{c}) small network, and (\textbf{d}) small network with layer norm. 
    Each line is the average of 30 runs; the shaded regions correspond to the standard error.
    Selective weight reinitialization maintained plasticity in all four settings, whereas ReDo and continual backpropagation were most successful in the large network setting.
    }
    \vspace{-10pt}
    \label{fig:pmnist_four_settings}
\end{figure}

We employ four network architectures in the permuted MNIST problem to evaluate the effectiveness of reinitializing weights versus units.
First, we use the same network as in the previous section, which we refer to as the \textit{large network} setting.
Second, we add layer normalization to the large network, \textit{large network with layer norm} setting, to confirm or disprove the anecdotal observations we received.
Third, we use a smaller version of the large network with only ten units per layer, \textit{small network} setting, to assess the effectiveness of reinitializing units in networks with limited capacity.
Finally, we add layer normalization to the small network, \textit{small network with layer norm} setting, to test how the two settings interact.
We use the online accuracy averaged over each task as the performance measure.

For each architecture, we present the performance of six learning systems in Figure \ref{fig:pmnist_four_settings}: 
the network architecture trained as is, labeled \textit{base system} in black;
the base system with L2-regularization, \textit{L2-regularization} in \pink; 
the L2-regularization system with parameter noise, \textit{shrink and perturb} in \vermilion;
the base system trained using continual backpropagation, \textit{continual backpropagation} in \orange; 
the base system trained using ReDo, \textit{ReDo} in \green; 
and the base system trained using selective weight reinitialization with gradient utility, threshold pruning, and resample reinitialization, \textit{selective weight reinitialization} in \blue. 
We used a grid search to tune the hyper-parameters of each learning system for each setting. 
We include more details about hyper-parameter selection in Appendix \ref{app:mnist_tuning} and present results with other settings of selective weight reinitialization in Appendix \ref{app:mnist_extended}.

Comparing the results with and without layer norm (left and right columns of Figure \ref{fig:pmnist_four_settings}, respectively), we confirm the anecdotal observations: reinitializing units is less effective at maintaining plasticity when the network employs layer norm. 
When using a large network, using continual backpropagation or ReDO resulted in stable performance without layer norm (Figure \ref{fig:pmnist_four_settings}a). 
However, when using layer norm with a large network (Figure \ref{fig:pmnist_four_settings}b), continual backpropagation did not maintain plasticity, and ReDo performed at a lower level than selective weight reinitialization.
In the small network, using both continual backpropagation and ReDo resulted in a drop in performance in the networks with and without layer norm (bottom row of Figure \ref{fig:pmnist_four_settings}); this effect was more severe when using layer norm (Figure \ref{fig:pmnist_four_settings}d).
On the other hand, selective weight reinitialization yielded stable performance in both networks with and without layer normalization.

The results with layer norm (Figures \ref{fig:pmnist_four_settings}b and \ref{fig:pmnist_four_settings}d) may appear surprising since layer norm has been reported to be useful for maintaining plasticity \citep{lyle2024normalization}.
However, \citeauthor{lyle2024normalization} (\citeyear{lyle2024normalization}) showed that layer norm prevents plasticity loss only when using an additional weight projection step.
When using layer norm alone, previous work has shown that the corresponding network loses plasticity over time \citep{kumar2024maintaining}.
Thus, our layer norm results are consistent with previous work. 

We confirm our initial intuition that reinitializing units is less effective than reinitializing weights in networks with a small number of units.
In the small network without layer norm setting (Figure \ref{fig:pmnist_four_settings}c), the performance of continual backpropagation and ReDo decreased after the first few tasks but stabilized soon after.
In contrast, the performances of continual backpropagation and ReDo were stable in the large network without layer norm (Figure \ref{fig:pmnist_four_settings}a).
When using layer norm and a small network, both algorithms experienced a steep decrease in performance compared to the large network with layer norm setting (Figures \ref{fig:pmnist_four_settings}b and \ref{fig:pmnist_four_settings}d, respectively). 

\textbf{Takeaways.} 
We found two settings where reinitializing units was less effective at maintaining plasticity than reinitializing weights: when the network uses layer norm and when it has a small number of units.
The first setting is particularly relevant for modern applications because layer norm has become the standard approach for normalizing activations in transformer architectures \citep{vaswani2017attention, devlin2018bert, dosovitskiy2021an}, the architectures responsible for the success of large language models.
The second setting is relevant for continual learning.
If one subscribes to the big world hypothesis \citep{javed2024the}, which poses that a learning system should be orders of magnitude smaller than the world they are learning about, then one can no longer rely on the size of the network to design successful learning algorithms.
Working under the big world hypothesis, reinitializing weights is more effective than reinitializing units because it does not rely on many units to maintain plasticity.
Finally, although L2-regularization and shrink and perturb are strong baselines in this problem, selective weight reinitialization achieved equal or better performance than L2-regularization in three of the four settings and better performance than shrink and perturb when using layer normalization.

Selective weight reinitialization maintained plasticity in all four settings and had the highest performance among all the systems when using layer normalization.
The stable performance of selective weight reinitialization can also be observed when using different optimizers.
We show the results for those experiments in Appendix \ref{app:mnist_extended}.

\section{Reinitializing weights vs reinitializing units for maintaining plasticity in Vision Transformers}
\label{sec:lop_in_vit}

We proceed to a large-scale demonstration using real-world data.
This section aims to assess the effectiveness of reinitializing weights in maintaining plasticity for more complex problems using a modern architecture.
Our secondary goal is to determine if reinitializing weights still shows an advantage over reinitializing units for maintaining plasticity in more complex problems.

We use the class incremental CIFAR-100 problem studied in by \citeauthor{chaudhry2018reimannian} (\citeyear{chaudhry2018reimannian}) and \citeauthor{dohare2024loss} (\citeyear{dohare2024loss}).
In this problem, networks are trained on an increasing number of classes from the CIFAR-100 dataset \citep{krizhevsky2009}, which comprises 100 classes with 500 training images and 100 test images per class.
In the first task, the network is trained to predict five classes.
After 100 epochs, the number of classes in the dataset increases by five, and a new task begins.
This process continues until the dataset contains all 100 classes, resulting in 20 tasks and 2,000 training epochs.

To isolate the loss of plasticity effect, we implement measures for preventing forgetting and overfitting.
To prevent forgetting, the network is constantly retrained on old classes; new classes are added to the dataset, but old classes are never removed from it.
To prevent overfitting, we use image transformations and early stopping.
For image transformation, we use random cropping with a padding of 4 pixels on each side of the image, random horizontal flipping with 0.5 probability, and random rotation between 0 and 15 degrees. 
To implement early stopping, we measure the network's accuracy after each epoch on a validation dataset of 50 images per class taken from the training set; at the start of each new task, we reset the network's parameters to those that resulted in the highest validation accuracy. 
The measure of performance is the highest accuracy achieved on the test set of each task.
Note that the network is never trained on test or validation sets, so our measure of performance is a measure of generalization. 

We utilize a vision transformer to generate predictions for images in the CIFAR-100 dataset.
We train the network using stochastic gradient descent with a momentum of 0.9.
We employ a linear learning rate schedule that increases the learning rate to 0.01 for the first 30 epochs and then decreases it to zero during the last 70 epochs.
We restart the learning rate schedule at the start of each new task.
We train the architecture using dropout, layer normalization, and L2-regularization.

\begin{figure}[t]
    \centering
    \includegraphics[width=\linewidth]{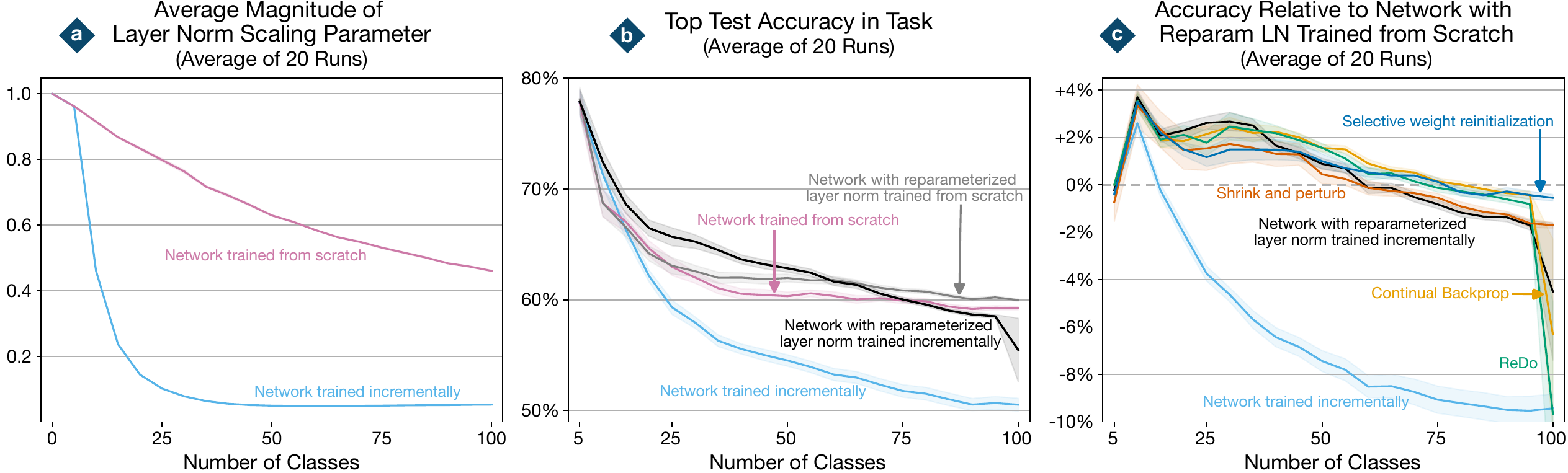}
    \vspace{-20pt}
    \caption{
    (\textbf{a}) Average magnitude of the scaling parameter in layer norm, (\textbf{b}) highest accuracy per task, and (\textbf{c}) accuracy relative to the network with reparameterized layer norm trained from scratch.
    Each line is the average of 20 runs; the shaded regions correspond to the standard error.
    Only selective weight reinitialization mitigated plasticity loss and maintained a stable performance.
    }
    \label{fig:vit_experiment}
    \vspace{-10pt}
\end{figure}

When first running this experiment, we noticed a pathological case in the layer norm modules of the architecture. 
As a reminder, layer normalization performs the operation:
$y = \frac{x - \bar{x}}{s_x + \epsilon} \cdot \gamma + \beta$,
where $x$ is a unit in the network, $\bar{x}$ and $s_x$ are the sample average and sample standard deviation of the units in the layer, $\epsilon$ is a small number to prevent division by zero, and $\gamma$ and $\beta$ are learnable scaling and shifting parameters. 
In our experiment, the magnitude of the scaling parameter drastically decreased when training the network incrementally (\cyan line in Figure \ref{fig:vit_experiment}a).
This decrease in magnitude harms learning by severely shrinking the gradients flowing back from the layer norm module.
To alleviate this problem, we used a reparameterized version of layer norm that performs the operation:
$y = \frac{x -\bar{x}}{s_x + \epsilon} \cdot (1 + \gamma) + \beta.$

Employing the reparameterized layer norm led to an improvement in performance for the vision transformer trained incrementally.
Figure \ref{fig:vit_experiment}b shows the performance of the network trained incrementally and from scratch, with and without reparameterized layer norm. 
When not using reparameterized layer norm, the difference in accuracy on the last task between the network trained incrementally and the network trained from scratch is close to ten percent (\cyan \ and \pink \ lines in Figure \ref{fig:vit_experiment}b, respectively).
On the other hand, when using the reparameterized layer norm, the difference in accuracy is only four percent.
However, we also detected other learning instabilities, which caused one of the runs with reparameterized layer norm to diverge during the last task, which is reflected in the wide confidence interval in Figure \ref{fig:vit_experiment}b. 
We discuss these learning instabilities in more detail in Appendix \ref{app:vit_instability}.
Moreover, we discuss different settings of the systems with reparameterized and standard layer norm, as well as various ways of applying regularization to the network's parameters, in Appendix \ref{app:vit_tuning}. 

We compare the performance of shrink and perturb, continual backpropagation, ReDo, and selective weight reinitialization when combined with the vision transformer architecture using the reparameterized layer norm.
We apply continual backpropagation and ReDo only between fully-connected layers in the network, as extensions of these methods have yet to be proposed for attention layers. 
We apply shrink and perturb and selective weight reinitialization to all the parameters in the architectures.
For details on hyper-parameter selection, see Appendix \ref{app:vit_tuning}.

Figure \ref{fig:vit_experiment}c shows the performance difference between each learning system and the network with reparameterized layer norm trained from scratch on the same set of classes.
Lines above the zero dashed line indicate a positive transfer of knowledge, whereas lines below zero indicate loss of plasticity. 
All the reinitialization algorithms had similar performance during the first 19 tasks.
However, on the last task, ReDo and continual backpropagation suffered from similar learning instabilities as the ones observed in the network trained incrementally. 
On the other hand, selective weight reinitialization decreased the drop in performance relative to the network trained from scratch, and it did not suffer from any learning instabilities. 
Nevertheless, selective weight reinitialization did not fully maintain plasticity on the last couple of tasks; its performance lies below the dashed line in Figure \ref{fig:vit_experiment}c.
Finally, shrink and perturb did not prevent loss of plasticity, but it helped prevent the learning instabilities observed in the network trained incrementally.

\textbf{Takeaways.}
The results in vision transformers confirm that reinitializing weights is a viable scheme for maintaining plasticity in more complex architectures, although it is not entirely successful. 
Reinitializing units is also a viable scheme, but unfortunately, the learning system suffered from learning instabilities.
Finally, a key to the success of the reinitialization algorithms in vision transformers was the use of reparameterized layer norm; without it, none of the algorithms were able to decrease the gap between the performance of the network trained from scratch and the network trained incrementally; we show those results in Appendix \ref{app:vit_instability}.

\section{Conclusion and future work}

In this paper, we present an algorithm for reinitializing weights to maintain plasticity, an approach that has remained unexplored in the literature on loss of plasticity.
Through comparisons in continual supervised learning, we uncovered two settings where reinitializing weights is more effective at maintaining plasticity than reinitializing units.
Moreover, the idea of reinitializing weights is easier to implement than reinitializing units since it does not have to account for the complex interconnections between the structures in the network, which is a difficulty also encountered in structural pruning \citep{fang2023depgraph}.
Finally, we demonstrated in a class-incremental problem that reinitializing weights partially maintains plasticity in larger architectures that employ many of the techniques used in modern applications. 

Selective weight reinitialization is easy to implement and highly modular.
This modularity enables easy modifications without altering the algorithm's functionality.
In this paper, we explored choices for the different settings of selective weight reinitialization commonly encountered in the literature. 
However, we expect further advances in utility functions, pruning functions, and reinitialization methods will lead to even better performance. 

One aspect of the continual learning problem that we did not consider in this paper was the issue of forgetting.
Designing algorithms that can learn without catastrophically forgetting is also an important endeavour in continual learning \citep{mccloskey1989, french1999}.
There is no evidence to suggest that the reinitialization algorithms presented in this paper can prevent forgetting.
It is likely that they exacerbate forgetting, as they erase parts of the network that are no longer useful and prioritize learning in the current task. 
Nevertheless, reinitialization algorithms could be adapted to avoid erasing past information. 
For example, the utility function could be computed using observations from previous and current tasks. 
Combined with a rehearsal method such as iCaRL \citep{rebuffi2017icarl}, this approach could potentially address both loss of plasticity and forgetting at once. 
That is beyond the scope of this paper, but it is an interesting avenue for future work.

Another limitation of our study is that we only considered the continuous supervised learning problem.
Loss of plasticity is also present in deep reinforcement learning, where both the input and target have non-stationary distributions \citep{nikishin2022primacy, nikishin2023deep}.
Another avenue for future work is to investigate whether our results can be extended to deep reinforcement learning. 

Lastly, as mentioned in the related work section, selective weight reinitialization shares characteristics with dynamic sparse training algorithms but entirely focuses on maintaining plasticity.
Adapting selective weight reinitialization to train a sparse network and maintain plasticity is possible.
The algorithm would be initialized with a sparse network, and, instead of reinitializing weights, it would prune active and regrow inactive connections in the same manner described in Section \ref{sec:swr}.
If implemented in a truly sparse fashion, this algorithm could result in fast learning while maintaining plasticity. 

\section*{Acknowledgements}
We gratefully acknowledge the Digital Research Alliance of Canada and Huawei Technologies Canada for providing the computational resources to carry out the experiments in this paper. 
We acknowledge funding from the Natural Sciences and Engineering Research Council (NSERC) of Canada, the Canada CIFAR AI Chairs Program, CIFAR and the Alberta Machine Intelligence Institute (Amii). 
We thank the members of the Reinforcement Learning and Artificial Intelligence (RLAI) laboratory for creating a stimulating and supportive research environment that made this work possible. 
Finally, we would like to thank Bram Grooten and Rohan Saha for the many stimulating discussions related to our algorithm and results.

\bibliography{collas2025_conference}
\bibliographystyle{collas2025_conference}

\newpage

\appendix

\section{Permuted MNIST Extended Results}   
\label{app:mnist_extended}                  

\begin{figure}[ht]
    \centering
    \includegraphics[width=\linewidth]{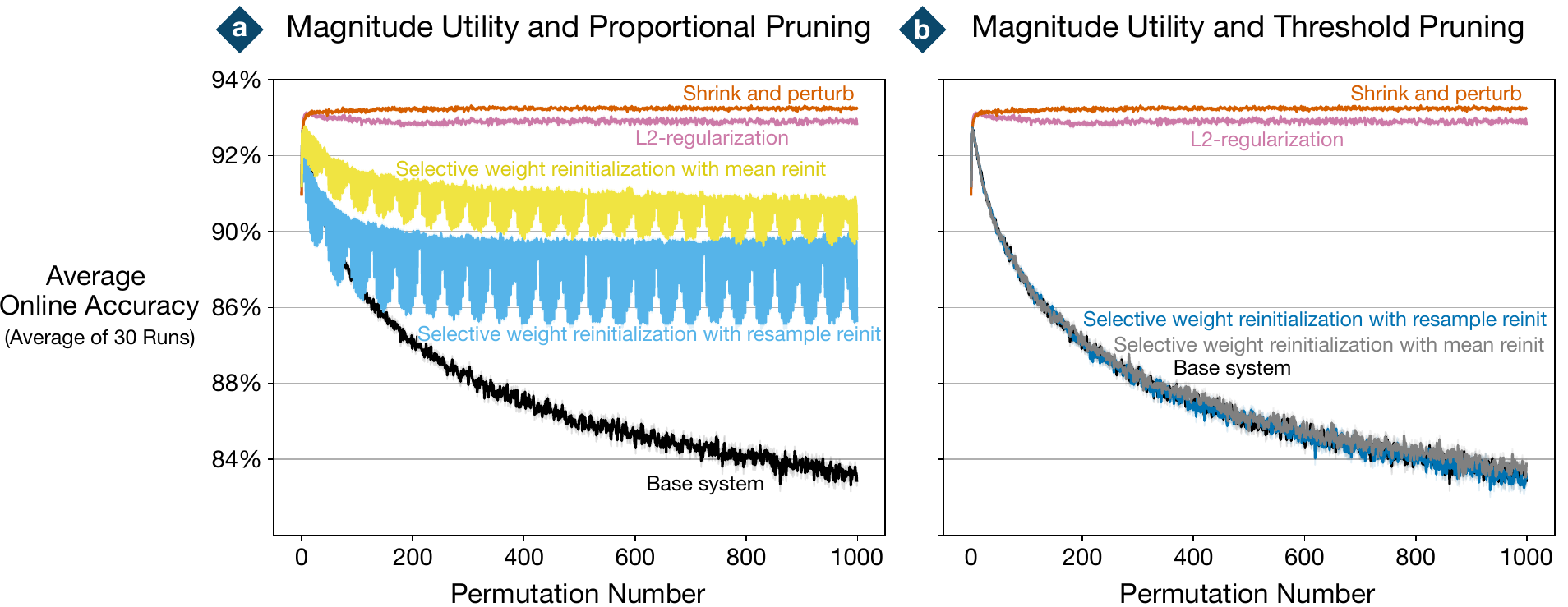}
    \vspace{-20pt}
    \caption{
    Average online accuracy of selective weight reinitialization with magnitude utility in the large network setting with (\textbf{a}) proportional pruning and (\textbf{b}) threshold pruning.
    Each line is the average of 30 runs.
    Shaded regions correspond to the standard error. 
    }
    \label{fig:swr_magnitude_utility}
\end{figure}

\textbf{Selective weight reinitialization with magnitude utility.}
We omitted the results with magnitude utility in Section \ref{sec:swr} in the main text of the paper.
We include those results in Figure \ref{fig:swr_magnitude_utility}.
Selective weight reinitialization with magnitude utility was not very effective at maintaining plasticity. 
With proportional pruning, there was a slight improvement over the base system; however, a drop in performance still occurred.
Moreover, there was a lot of variability in performance due to the extremely high reinitialization factor (see Table \ref{tb:initial_assessment} for more details). 
There was no improvement in performance when using threshold pruning and magnitude utility (Figure \ref{fig:swr_magnitude_utility}b).

\begin{figure}[!ht]
    \centering
    \includegraphics[width=0.7\linewidth]{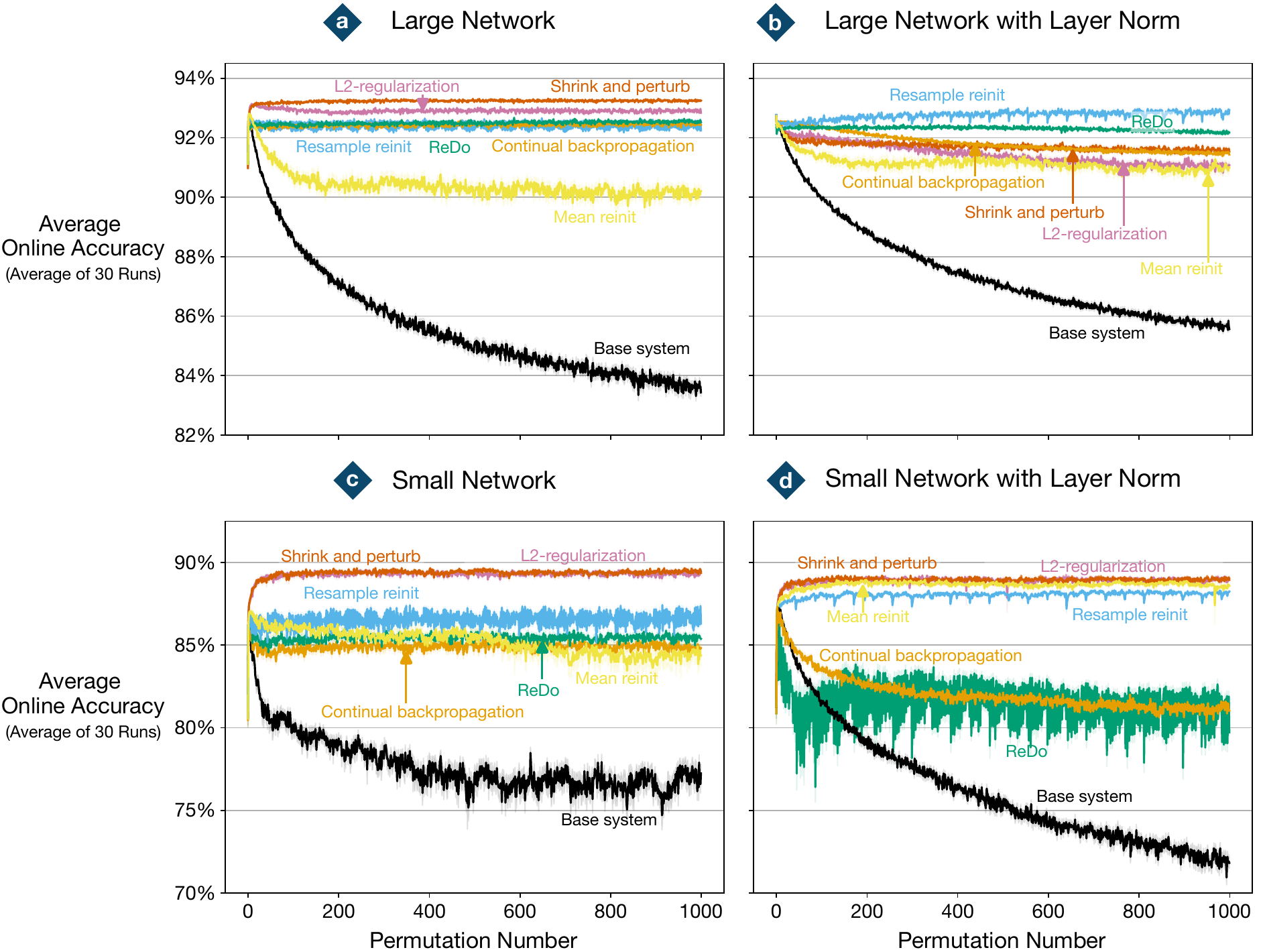}
    \caption{
    Average online accuracy of selective weight reinitialization with gradient utility and proportional pruning in (\textbf{a}) the large network setting, (\textbf{b}) the large network with layer norm setting, (\textbf{c}) the small network setting, and (\textbf{d}) the small network with layer norm setting.
    Each line is the average of 30 runs.
    Shaded regions correspond to the standard error.
    }
    \label{fig:four_settings_proportion}
\end{figure}

\begin{figure}[!ht]
    \centering
    \includegraphics[width=0.7\linewidth]{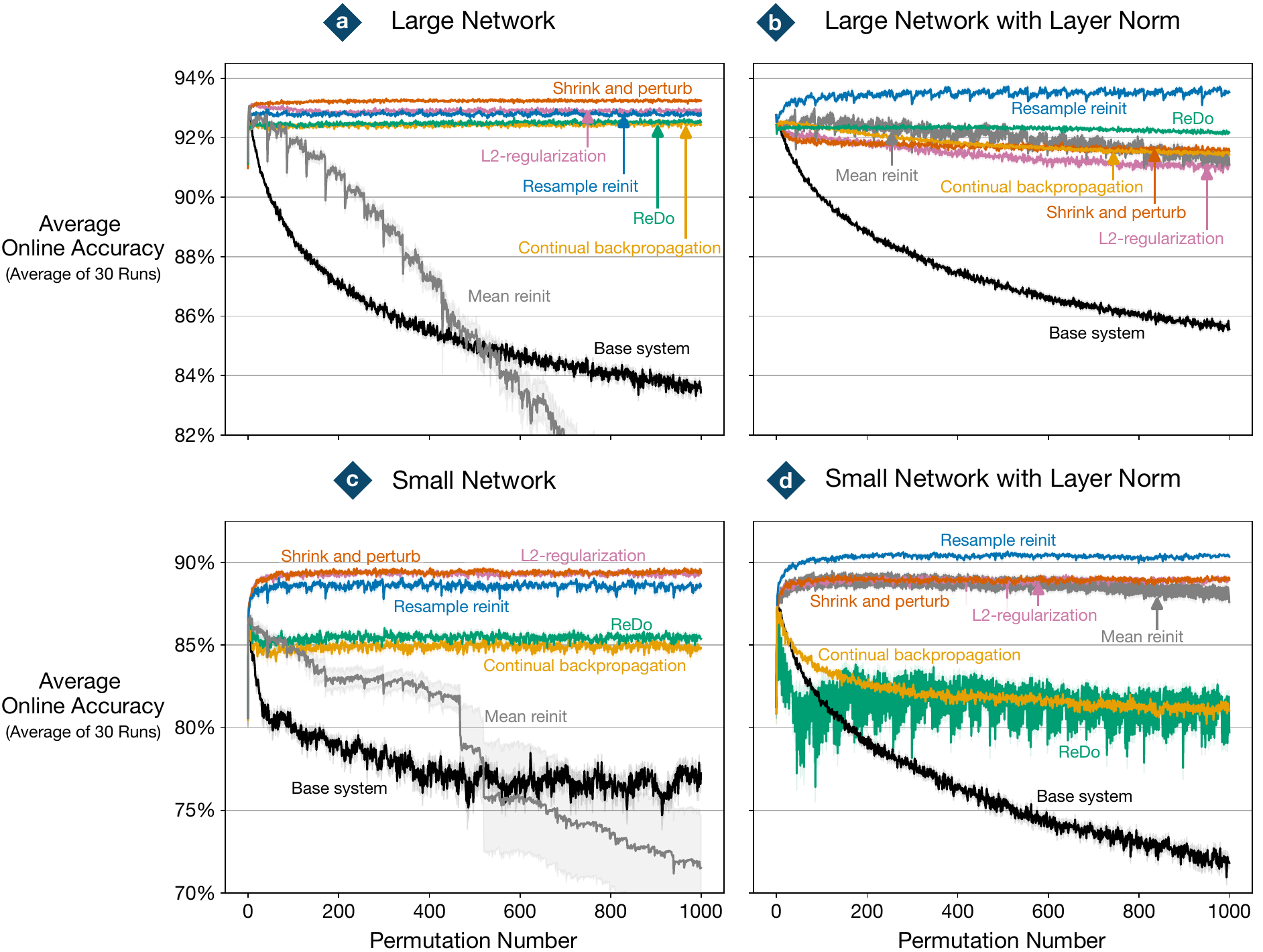}
    \caption{
    Average online accuracy of selective weight reinitialization with gradient utility and threshold pruning in (\textbf{a}) the large network setting, (\textbf{b}) the large network with layer norm setting, (\textbf{c}) the small network setting, and (\textbf{d}) the small network with layer norm setting.
    Each line is the average of 30 runs.
    Shaded regions correspond to the standard error.
    }
    \label{fig:four_settings_threshold}
\end{figure}

\textbf{Extended results for selective weight reinitialization with gradient utility.}
In Section \ref{sec:permuted_mnist}, we presented results using selective weight reinitialization in four different settings in Permuted MNIST. 
We chose to present the results using threshold pruning, gradient utility, and resample reinitialization because this setting proved to be the most effective in maintaining plasticity.
However, other settings of selective weight reinitialization also show some success.
We include results with proportional pruning and gradient utility in Figure \ref{fig:four_settings_proportion} and results with threshold pruning and gradient utility in Figure \ref{fig:four_settings_threshold}.

In the case of proportional pruning, we see similar results to those with threshold pruning. 
Selective weight reinitialization with resample reinitialization had stable performance in all four settings.
Moreover, in all but one setting, resample reinitialization resulted in higher performance than ReDo and continual backpropagation.
However, resample reinitialization only surpassed the performance of shrink and perturb and L2-regularization in the large network with layer norm setting (Figure \ref{fig:four_settings_proportion}b).
Mean reinitialization resulted in decreasing performance in all but one setting.
However, it performed higher than resample reinitialization in the small network with layer norm setting (Figure \ref{fig:four_settings_proportion}d), suggesting that resampling reinitialization is not always the best choice of reinitialization method. 

In the case of threshold pruning, we found little success with mean reinitialization (Figure \ref{fig:four_settings_threshold}a and c). 
In all four settings, mean reinitialization resulted in a decrease in performance as the number of tasks increased. 
Overall, mean reinitialization was rarely a sensible choice of reinitialization method. 

\newpage

\textbf{Reinitializing weights vs units using different optimizers.}
All the experiments in Sections \ref{sec:swr} and \ref{sec:permuted_mnist} were done using stochastic gradient descent (SGD).
This begs the question: does selective weight reinitialization work with different optimizers?
To answer this question, we reused the large network setting, using SGDW with momentum and AdamW \citep{loshchilov2018decoupled} to optimize the network's parameters. 
For SGDW, we explored different values of the momentum term and the step size parameter.
For AdamW, we tested various values for the step size parameter and the moving average factors, $\beta_1$ and $\beta_2$. 
For more details on hyper-parameter tuning, see the next section.

\begin{figure}[h!]
    \centering
    \includegraphics[width=\linewidth]{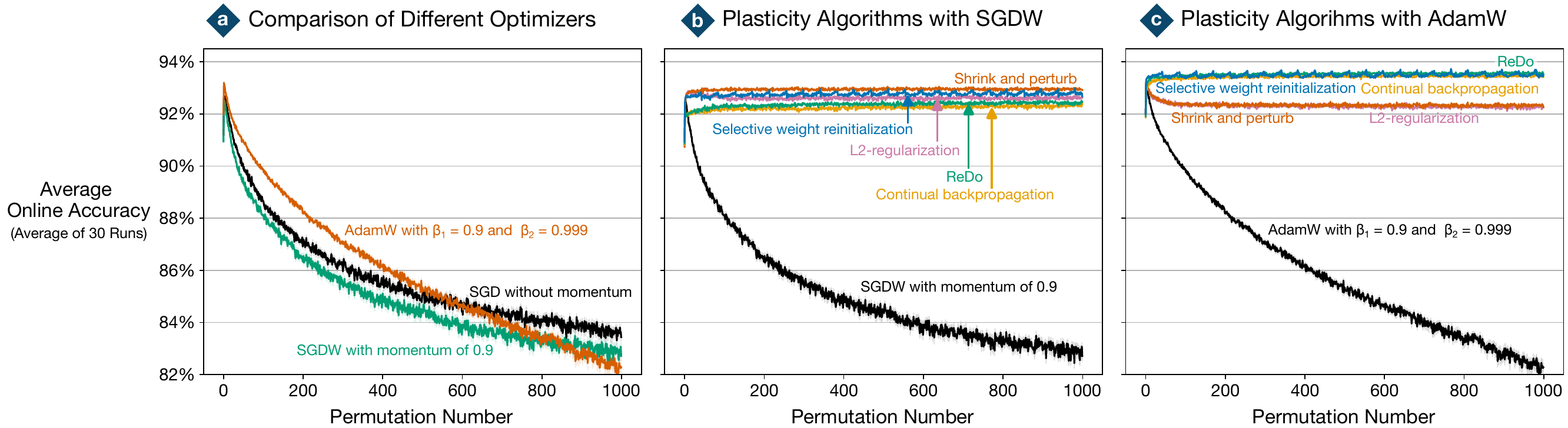}
    \caption{
    Comparison with different optimizers in the large network setting.
    (\textbf{a}) Comparison between SGD, SGDW with momentum, and AdamW. 
    (\textbf{b}) SGDW with momentum augmented by various algorithms to prevent plasticity loss. 
    (\textbf{c}) AdamW augmented by various algorithms to prevent plasticity loss. 
    Each line is the average of 30 runs.
    Shaded regions correspond to the standard error.
    }
    \label{fig:optim_comparisons}
\end{figure}

Once we tuned the hyper-parameters for SGDW and AdamW, we added L2-regularization, shrink and perturb, continual backpropagation, ReDo, and selective weight reinitialization to each optimizer, in the same manner as in Section \ref{sec:permuted_mnist}.
In case of selective weight reinitialization, we used threshold pruning, gradient utility, and resample reinitialization.
We show the results in Figure \ref{fig:optim_comparisons}.

All three different optimizers resulted in plasticity loss (Figure \ref{fig:optim_comparisons}a).
Nevertheless, selective weight reinitialization was successful at maintaining plasticity regardless of the choice of optimizer.
When using SGDW with momentum, selective weight reinitialization achieved the second-highest performance, outperforming ReDo and continual backpropagation. 
When using AdamW, selective weight reinitialization, ReDo, and continual backpropagation achieved the same level of performance, and all performed higher than L2-regularization and shrink and perturb. 
Both L2-regularization and shrink and perturb had a slight decrease in performance when using AdamW. 

Together with the results in Section \ref{sec:permuted_mnist}, selective weight reinitialization with threshold pruning, gradient utility, and resample reinitialization reliably maintained plasticity in the Permuted MNIST problem across six different settings. 

\newpage

\section{Hyper-parameter tuning in permuted MNIST}
\label{app:mnist_tuning}

For the experiments in Permuted MNIST presented in Sections \ref{sec:swr} and \ref{sec:permuted_mnist}, we tuned the hyper-parameters of each learning system using a grid search with ten runs per hyper-parameter combination. 
After the search, we selected the hyper-parameter values that resulted in the highest average online accuracy throughout the training period, i.e., the area under the curve.
We tuned the step size parameter, $\alpha$, for the base systems.
For the learning systems using L2-regularization, we tuned the regularization factor, $\lambda$.
For continual backpropagation, we tuned the replacement rate, $rr$, and maturity threshold, $mt$.
For Redo, we tuned the reinitialization frequency, $rf$, and reinitialization threshold, $rt$.
For selective weight reinitialization, we tuned the reinitialization frequency, $\tau$, and reinitialization factor, $k$.
For shrink and perturb, we tuned the variance of the noise, $\sigma^2$, and used the same regularization factor as the L2-regularization learning systems. 
The learning systems using L2-regularization, shrink and perturb, continual backpropagation, ReDo, and selective weight reinitialization used the same step size as the base systems in the corresponding setting. 
Finally, for the layer normalization settings, we compared layer norm before and after the activation. 
Using layer norm after the activation resulted in higher average online accuracy in small and large networks.

For the results in Appendix \ref{app:mnist_extended}, we also used a grid search as described above.
For SGDW with momentum, we tuned the step size parameter, $\alpha$, and the momentum term, $m$.
For AdamW, we tuned the step size parameter, $\alpha$, the exponential moving average factor for the first moment, $\beta_1$, and the exponential moving average for the second moment, $\beta_2$. 
For $\epsilon$ in AdamW, we used the default value of $10^{-8}$.
We augmented the SGDW with momentum and the AdamW systems with L2-regularization, shrink and perturb, continual backpropagation, ReDo, or selective weight reinitialization with threshold pruning, gradient utility, and resample reinitialization. 
We tuned the parameters of each of these learning systems in the same fashion described in the previous paragraph. 

Table \ref{tb:initial_assessment} shows the hyper-parameter values tested for each algorithm in the initial assessment presented in Section \ref{sec:swr}.
Underlined values correspond to the values used in the main text.
For selective weight reinitialization (SWR), we labelled best hyper-parameter values \textbf{GR} for gradient utility with resample reinitialization, \textbf{GM} for gradient utility with mean reinitialization, \textbf{MR} for magnitude utility with resample reinitialization, and \textbf{MM} for magnitude utility with mean reinitialization. 
Table \ref{tb:pmnist_large_hp} and \ref{tb:pmnist_small_hp} show the hyper-parameter values tested for the four different settings in Section \ref{sec:permuted_mnist}. 
Finally, Table \ref{tb:optimizer_comparisons_hp} shows the hyper-parameter values tested for the optimizer comparisons presented in the previous section. 

\begin{table}[h]
\caption{
Hyper-parameter values used for the initial assessment of selective weight reinitialization.
}
\label{tb:initial_assessment}
\begin{center}
\begin{tabular}{l c}
Base system      
& $\alpha \in \{0.1, \underline{0.05}, 0.01, 0.005, 0.001, 0.0005\}$ \\[1mm]
\hdashline \\[-1.7mm]
L2-regularization    
& $\lambda \in \{ 10^{-3}/\alpha, \underline{10^{-4}/\alpha}, 10^{-5}/\alpha, 10^{-6}/\alpha, 10^{-7}/\alpha, 10^{-8}/\alpha\}$ \\[1mm]
\hdashline \\[-1.7mm]
Shrink and perturb
& $\sigma^2 \in \{10^{-4}, 10^{-5}, 10^{-6}, \underline{10^{-7}}, 10^{-8}\}$ \\[1mm]
\hdashline \\[-1.7mm]
\multirow{2}{10em}{SWR Proportional Pruning}             
& $\tau \in \{2^8, 2^9 \ \textbf{(GM)}, 2^{10}, 2^{11}, 2^{12} \ \textbf{(GR)}, 2^{13} \ \textbf{(MM, MR)}\}$ \\
& $k \in \{0.01, 0.05, 0.1, 0.2 \ \textbf{(GM)}, 0.4 \ \textbf{(GR)}, 0.8 \ \textbf{(MM, MR)} \}$ \\[1mm]
\hdashline \\[-1.7mm]
\multirow{2}{10em}{SWR Threshold Pruning}             
& $\tau \in \{2^8, 2^9, 2^{10} \ \textbf{(MM)}, 2^{11} \ \textbf{(MR, GM, GR)}, 2^{12}, 2^{13}\}$ \\
& $k \in \{10^{-6}, 10^{-5} \ \textbf{(GR)}, 10^{-4} \ \textbf{(MR)}, 10^{-3} \ \textbf{(GM)}, 10^{-2}, 10^{-1} \ \textbf{(MM)}\}$ \\[1mm]
\end{tabular}
\end{center}
\end{table}

\newpage

\begin{table}[h]
\caption{Hyper-parameter values used in the large network settings in Permuted MNIST.}
\label{tb:pmnist_large_hp}
\begin{center}
\begin{tabular}{l c}
\multicolumn{2}{c}{\bf Large Network Setting} 
\\[1mm] \hline \\[-1.7mm]
Base system      
& $\alpha \in \{0.1, \underline{0.05}, 0.01, 0.005, 0.001, 0.0005\}$ \\[1mm]
\hdashline \\[-1.7mm]
L2-regularization    
& $\lambda \in \{ 10^{-3}/\alpha, \underline{10^{-4}/\alpha}, 10^{-5}/\alpha, 10^{-6}/\alpha, 10^{-7}/\alpha, 10^{-8}/\alpha\}$ \\[1mm]
\hdashline \\[-1.7mm]
Shrink and perturb
& $\sigma^2 \in \{10^{-4}, 10^{-5}, 10^{-6}, \underline{10^{-7}}, 10^{-8}\}$ \\[1mm]
\hdashline \\[-1.7mm]
\multirow{2}{10em}{Continual backpropagation}             
& $rr \in \{10^{-1}, 10^{-2}, 10^{-3}, \underline{10^{-4}}, {1\times 10^{-5}}\}$ \\
& $mt \in \{ 1, 5, 10, 50, 100, \underline{500}\}$ \\[1mm]
\hdashline \\[-1.7mm]
\multirow{2}{10em}{ReDo}  
& $rf \in \{2^3, \underline{2^4}, 2^5, 2^6, 2^7, 2^8\}$ \\
& $rt \in \{10^{-5}, \underline{10^{-4}}, 10^{-3}, 10^{-2}, 10^{-1}, 10^{0} \}$ \\[1mm]
\hdashline \\[-1.7mm]
\multirow{2}{10em}{SWR Proportional Pruning}             
& $\tau \in \{2^8, 2^9 \ \textbf{(GM)}, 2^{10}, 2^{11}, 2^{12} \ \textbf{(GR)}, 2^{13} \}$ \\
& $k \in \{0.01, 0.05, 0.1, 0.2 \ \textbf{(GM)}, 0.4 \ \textbf{(GR)}, 0.8 \}$ \\[1mm]
\hdashline \\[-1.7mm]
\multirow{2}{10em}{SWR Threshold Pruning}             
& $\tau \in \{2^8, 2^9, 2^{10}, 2^{11} \ \textbf{(GM, GR)}, 2^{12}, 2^{13}\}$ \\
& $k \in \{10^{-6}, 10^{-5} \ \textbf{(GR)}, 10^{-4}, 10^{-3} \ \textbf{(GM)}, 10^{-2}, 10^{-1} \}$ \\[1mm]
\hline \\[-1.7mm]
\multicolumn{2}{c}{\bf Large Network with Layer Norm Setting}
\\[1mm] \hline \\[-1.7mm]
Base system      
& $\alpha \in \{0.5, \underline{0.1}, 0.05, 0.01, 0.005, 0.001, 0.0005\}$ \\[1mm]
\hdashline \\[-1.7mm]
L2-regularization    
& $\lambda \in \{ 10^{-3}/\alpha, 10^{-4}/\alpha, \underline{10^{-5}/\alpha}, 10^{-6}/\alpha, 10^{-7}/\alpha, 10^{-8}/\alpha\}$ \\[1mm]
\hdashline \\[-1.7mm]
Shrink and perturb
& $\sigma^2 \in \{10^{-4}, 10^{-5}, 10^{-6}, \underline{10^{-7}}, 10^{-8}\}$ \\[1mm]
\hdashline \\[-1.7mm]
\multirow{2}{10em}{Continual backpropagation}  
& $rr \in \{ 10^{-1}, \underline{10^{-2}}, {10^{-3}}, 10^{-4}, 10^{-5} \}$ \\
& $mt \in \{\underline{1}, 5, 10, 50, 100, 500\}$ \\[1mm]
\hdashline \\[-1.7mm]
\multirow{2}{10em}{ReDo}  
& $rf \in \{ \underline{2^3}, 2^4, 2^5, 2^6, 2^7 \}$ \\
& $rt \in \{ 10^{-4}, 10^{-3}, 10^{-2}, \underline{10^{-1}}, 10^0 \}$ \\[1mm]
\hdashline \\[-1.7mm]
\multirow{2}{10em}{SWR Proportional Pruning}             
& $\tau \in \{2^8, 2^9 \ \textbf{(GM)}, 2^{10}, 2^{11} \ \textbf{(GR)}, 2^{12} \}$ \\
& $k \in \{0.05 \ \textbf{(GM)}, 0.1 \ \textbf{(GR)}, 0.2, 0.4, 0.8 \}$ \\[1mm]
\hdashline \\[-1.7mm]
\multirow{2}{10em}{SWR Threshold Pruning}             
& $\tau \in \{ 2^8, 2^9, 2^{10}, 2^{11} \ \textbf{(GR)}, 2^{12} \ \textbf{(GM)} \}$ \\
& $k \in \{ 10^{-6}, 10^{-5} \ \textbf{(GR)}, 10^{-4}, 10^{-3} \ \textbf{(GM)}, 10^{-2} \}$ \\[1mm]
\end{tabular}
\end{center}
\end{table}

\begin{table}
\caption{Hyper-parameter values used in each of the small network settings in Permuted MNIST.}
\label{tb:pmnist_small_hp}
\begin{center}
\begin{tabular}{l c}
\hline \\[-1.7mm]
\multicolumn{2}{c}{\bf Small Network Setting}
\\[1mm] \hline \\[-1.7mm]
Base system      
& $\alpha \in \{0.1, \underline{0.05}, 0.01, 0.005, 0.001, 0.0005\}$ \\[1mm]
\hdashline \\[-1.7mm]
L2-regularization    
& $\lambda \in \{ 10^{-3}/\alpha, \underline{10^{-4}/\alpha}, 10^{-5}/\alpha, 10^{-6}/\alpha, 10^{-7}/\alpha, 10^{-8}/\alpha\}$ \\[1mm]
\hdashline \\[-1.7mm]
Shrink and perturb
& $\sigma^2 \in \{ 10^{-4}, 10^{-5}, 10^{-6}, 10^{-7}, 10^{-8}, \underline{10^{-9}}, 10^{-10} \}$ \\[1mm]
\hdashline \\[-1.7mm]
\multirow{2}{10em}{Continual backpropagation}       
& $rr \in \{ 10^{-1}, 10^{-2}, \underline{10^{-3}}, 10^{-4}, 10^{-5} \}$ \\
& $mt \in \{ 1, \underline{5}, 10, 50, 100, 500 \}$ \\[1mm]
\hdashline \\[-1.7mm]
\multirow{2}{10em}{ReDo}  
& $rf \in \{ \underline{2^3}, 2^4, 2^5, 2^6, 2^7 \}$ \\
& $rt \in \{ 10^{-5}, 10^{-4}, 10^{-3}, \underline{10^{-2}}, 10^{-1} \}$ \\[1mm]
\hdashline \\[-1.7mm]
\multirow{2}{10em}{SWR Proportional Pruning}             
& $\tau \in \{ 2^8, 2^9 \ \textbf{(GM)}, 2^{10}, 2^{11}, 2^{12} \ \textbf{(GR)} \}$ \\
& $k \in \{ 0.01, 0.05, 0.1 \ \textbf{(GM)}, 0.2, 0.4 \ \textbf{(GR)} \}$ \\[1mm]
\hdashline \\[-1.7mm]
\multirow{2}{10em}{SWR Threshold Pruning}             
& $\tau \in \{ 2^8, 2^9, 2^{10}, 2^{11}  \ \textbf{(GM, GR)}, 2^{12} \}$ \\
& $k \in \{ 10^{-6} \ \textbf{(GR)}, 10^{-5}, 10^{-4}, 10^{-3} \ \textbf{(GM)}, 10^{-2} \}$ \\[1mm]
\hline \\[-1.7mm]
\multicolumn{2}{c}{\bf Small Network with Layer Norm Setting}
\\[1mm] \hline \\[-1.7mm]
Base system      
& $\alpha \in \{0.5, \underline{0.1}, 0.05, 0.01, 0.005, 0.001, 0.0005\}$ \\[1mm]
\hdashline \\[-1.7mm]
L2-regularization    
& $\lambda \in \{ 10^{-3}/\alpha, \underline{10^{-4}/\alpha}, 10^{-5}/\alpha, 10^{-6}/\alpha, 10^{-7}/\alpha, 10^{-8}/\alpha \}$ \\[1mm]
\hdashline \\[-1.7mm]
Shrink and perturb
& $\sigma^2 \in \{ 10^{-4}, 10^{-5}, 10^{-6}, 10^{-7}, 10^{-8}, 10^{-9}, \underline{10^{-10}}, 10^{-11} \}$ \\[1mm]
\hdashline \\[-1.7mm]
\multirow{2}{10em}{Continual backpropagation}             
& $rr \in \{ 10^{-2}, 10^{-3}, 10^{-4}, \underline{10^{-5}}, 10^{-6} \}$ \\
& $mt \in \{ \underline{1}, 5, 10, 50, 100, 500 \}$ \\[1mm]
\hdashline \\[-1.7mm]
\multirow{2}{10em}{ReDo}  
& $rf \in \{ 2^9, 2^{10}, 2^{11}, 2^{12}, \underline{2^{13}} \}$ \\
& $rt \in \{ 10^{-4}, 10^{-3}, 10^{-2}, 10^{-1}, \underline{10^0} \}$ \\[1mm]
\hdashline \\[-1.7mm]
\multirow{2}{10em}{SWR Proportional Pruning}             
& $\tau \in \{ 2^8, 2^9 \ \textbf{(GM)}, 2^{10}, 2^{11}, 2^{12} \ \textbf{(GR)} \}$ \\
& $k \in \{ 0.01, 0.05, 0.1 \ \textbf{(GM)}, 0.2, 0.4 \ \textbf{(GR)}\}$ \\[1mm]
\hdashline \\[-1.7mm]
\multirow{2}{10em}{SWR Threshold Pruning}             
& $\tau \in \{ 2^8, 2^9, 2^{10}, 2^{11} \ \textbf{(GM, GR)}, 2^{12} \}$ \\
& $k \in \{ 10^{-6} \ \textbf{(GR)}, 10^{-5}, 10^{-4}, 10^{-3} \ \textbf{(GM)}, 10^{-2} \}$ \\[1mm]
\end{tabular}
\end{center}
\end{table}

\begin{table}
\caption{Hyper-parameter values used in the optimizers comparisons in Permuted MNIST.}
\label{tb:optimizer_comparisons_hp}
\begin{center}
\begin{tabular}{l c}
\hline \\[-1.7mm]
\multicolumn{2}{c}{\bf Comparisons using SGDW}
\\[1mm] \hline \\[-1.7mm]
\multirow{2}{10em}{SGDW with momentum}       
& $\alpha \in \{ 10^{-2}, \underline{5 \cdot 10^{-3}}, 10^{-3}, 5\cdot 10^{-4}, 10^{-4}, 5 \cdot 10^{-5}, 10^{-5}, 5 \cdot 10^{-6}, 10^{-6}, 5 \cdot 10^{-7}, 10^{-7} \}$ \\
& $m \in \{ \underline{0.9}, 0.99, 0.999 \}$ \\[1mm]
\hdashline \\[-1.7mm]
L2-regularization    
& $\lambda \in \{ 10^{-2}/\alpha, 10^{-3}/\alpha, \underline{10^{-4}/\alpha}, 10^{-5}/\alpha, 10^{-6}/\alpha \}$ \\[1mm]
\hdashline \\[-1.7mm]
Shrink and perturb
& $\sigma^2 \in \{ 10^{-4}, 10^{-5}, 10^{-6}, \underline{10^{-7}}, 10^{-8}, 10^{-9} \}$ \\[1mm]
\hdashline \\[-1.7mm]
\multirow{2}{10em}{Continual backpropagation}       
& $rr \in \{ 10^{-1}, 10^{-2}, \underline{10^{-3}}, 10^{-4}, 10^{-5} \}$ \\
& $mt \in \{ \underline{1}, 5, 10, 50, 100, 500 \}$ \\[1mm]
\hdashline \\[-1.7mm]
\multirow{2}{10em}{ReDo}  
& $rf \in \{ \underline{2^3}, 2^4, 2^5, 2^6, 2^7 \}$ \\
& $rt \in \{ 10^{-5}, 10^{-4}, \underline{10^{-3}}, 10^{-2}, 10^{-1} \}$ \\[1mm]
\hdashline \\[-1.7mm]
\multirow{2}{10em}{Selective weight reinitialization}             
& $\tau \in \{ 2^8, 2^9, 2^{10}, \underline{2^{11}}, 2^{12} \}$ \\
& $k \in \{ \underline{10^{-6}}, 10^{-5}, 10^{-4}, 10^{-3}, 10^{-2} \}$ \\[1mm]
\hline \\[-1.7mm]
\multicolumn{2}{c}{\bf Comparisons using AdamW}
\\[1mm] \hline \\[-1.7mm]
\multirow{3}{10em}{AdamW}             
& $\alpha \in \{ 10^{-2}, 5 \cdot 10^{-3}, 10^{-3}, \underline{5 \cdot 10^{-4}}, 10^{-4}, 5 \cdot 10^{-5} \}$ \\
& $\beta_1 \in \{ \underline{0.9}, 0.99, 0.999 \}$ \\
& $\beta_2 \in \{ 0.9, 0.99, \underline{0.999} \}$ \\[1mm]
\hdashline \\[-1.7mm]
L2-regularization    
& $\lambda \in \{ 10^{-2}/\alpha, 10^{-3}/\alpha, 10^{-4}/\alpha, \underline{10^{-5}/\alpha}, 10^{-6}/\alpha \}$ \\[1mm]
\hdashline \\[-1.7mm]
Shrink and perturb
& $\sigma^2 \in \{ 10^{-4}, 10^{-5}, 10^{-6}, 10^{-7}, 10^{-8}, \underline{10^{-9}}, 10^{-10} \}$ \\[1mm]
\hdashline \\[-1.7mm]
\multirow{2}{10em}{Continual backpropagation}             
& $rr \in \{ 10^{-1}, 10^{-2}, \underline{10^{-3}}, 10^{-4}, 10^{-5} \}$ \\
& $mt \in \{ \underline{1}, 5, 10, 50, 100, 500 \}$ \\[1mm]
\hdashline \\[-1.7mm]
\multirow{2}{10em}{ReDo}  
& $rf \in \{ 2^3, \underline{2^4}, 2^5, 2^6, 2^7 \}$ \\
& $rt \in \{ \underline{10^{-5}}, 10^{-4}, 10^{-3}, 10^{-2}, 10^{-1} \}$ \\[1mm]
\hdashline \\[-1.7mm]
\multirow{2}{10em}{Selective weight reinitialization}             
& $\tau \in \{ 2^8, 2^9, 2^{10}, \underline{2^{11}}, 2^{12} \}$ \\
& $k \in \{ \underline{10^{-6}}, 10^{-5}, 10^{-4}, 10^{-3}, 10^{-2} \}$ \\[1mm]
\hdashline \\[-1.7mm]
\end{tabular}
\end{center}
\end{table}

\newpage
\newpage

\textbf{Hyper-parameter sensitivity.}
The performance of each algorithm presented in Sections \ref{sec:swr} and \ref{sec:permuted_mnist} and Appendix \ref{app:mnist_extended} is highly dependent on the choice of hyper-parameters.
Here we present sensitivity curves for each reinitialization algorithm.
Curves with flat regions suggest that the algorithm behaves similarly across multiple hyper-parameter settings, implying that it is not very sensitive to hyper-parameter choices.
Flat regions near the optimal hyper-parameter values indicate that it is not difficult to find hyper-parameter values that yield high performance. 

\begin{figure}[h!]
    \centering
    \includegraphics[width=\linewidth]{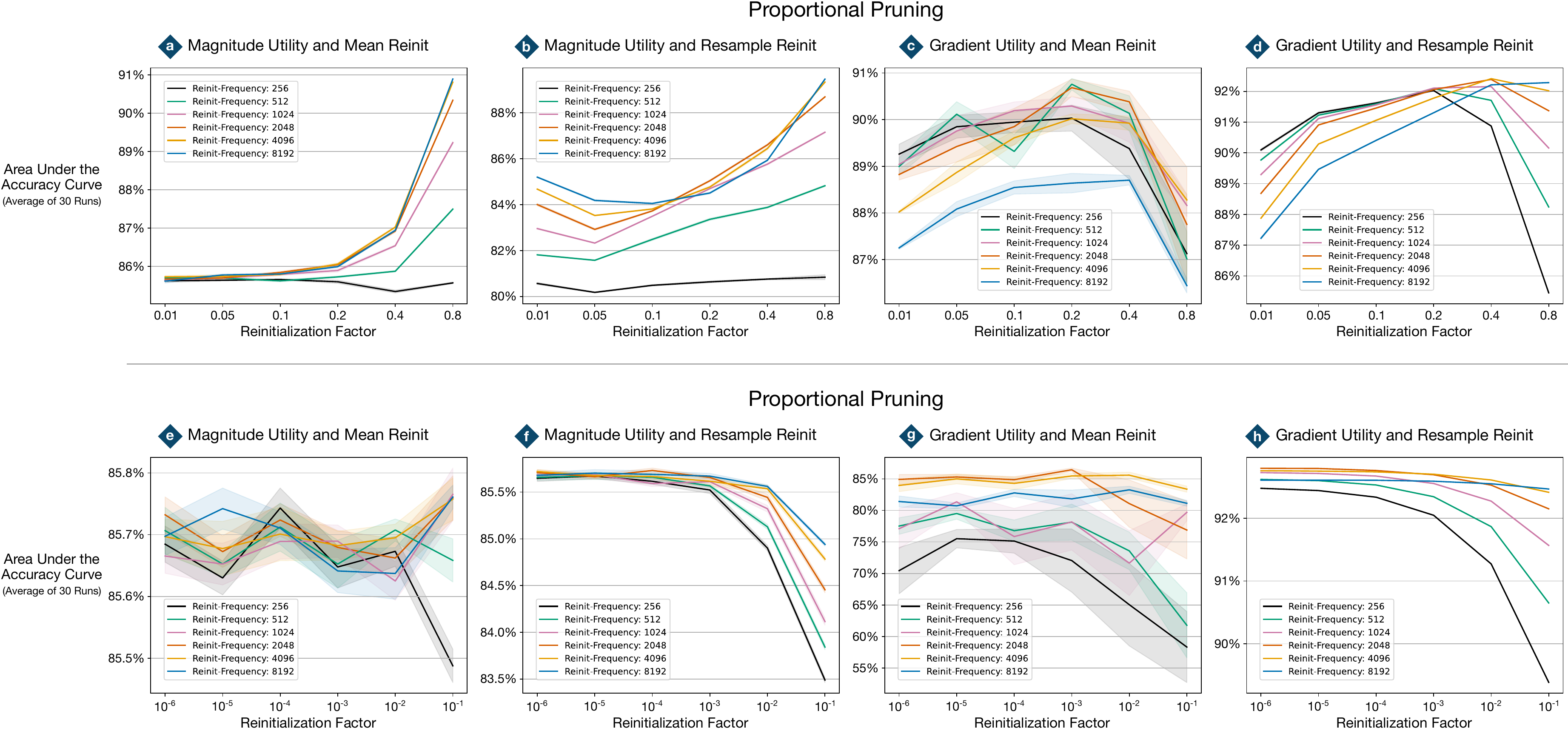}
    \caption{
    Sensitivity curve for selective weight reinitialization with proportional and threshold pruning.
    Each line is the average of 10 runs.
    Shaded regions correspond to the standard error.
    }
    \label{fig:initial_assessment_sensitivity}
\end{figure}

In Figure \ref{fig:initial_assessment_sensitivity}, we show the sensitivity curves for the initial assessment presented in Section \ref{sec:swr} and Appendix \ref{app:mnist_extended}. 
Selective weight reinitialization with proportional pruning was very sensitive to hyper-parameter settings, resulting in several peaks in the sensitivity curve. 
On the other hand, the sensitivity curve of threshold pruning and resample reinitialization (Figures \ref{fig:initial_assessment_sensitivity}f and \ref{fig:initial_assessment_sensitivity}h) shows flat regions near the highest point.

\begin{figure}[h!]
    \centering
    \includegraphics[width=0.72\linewidth]{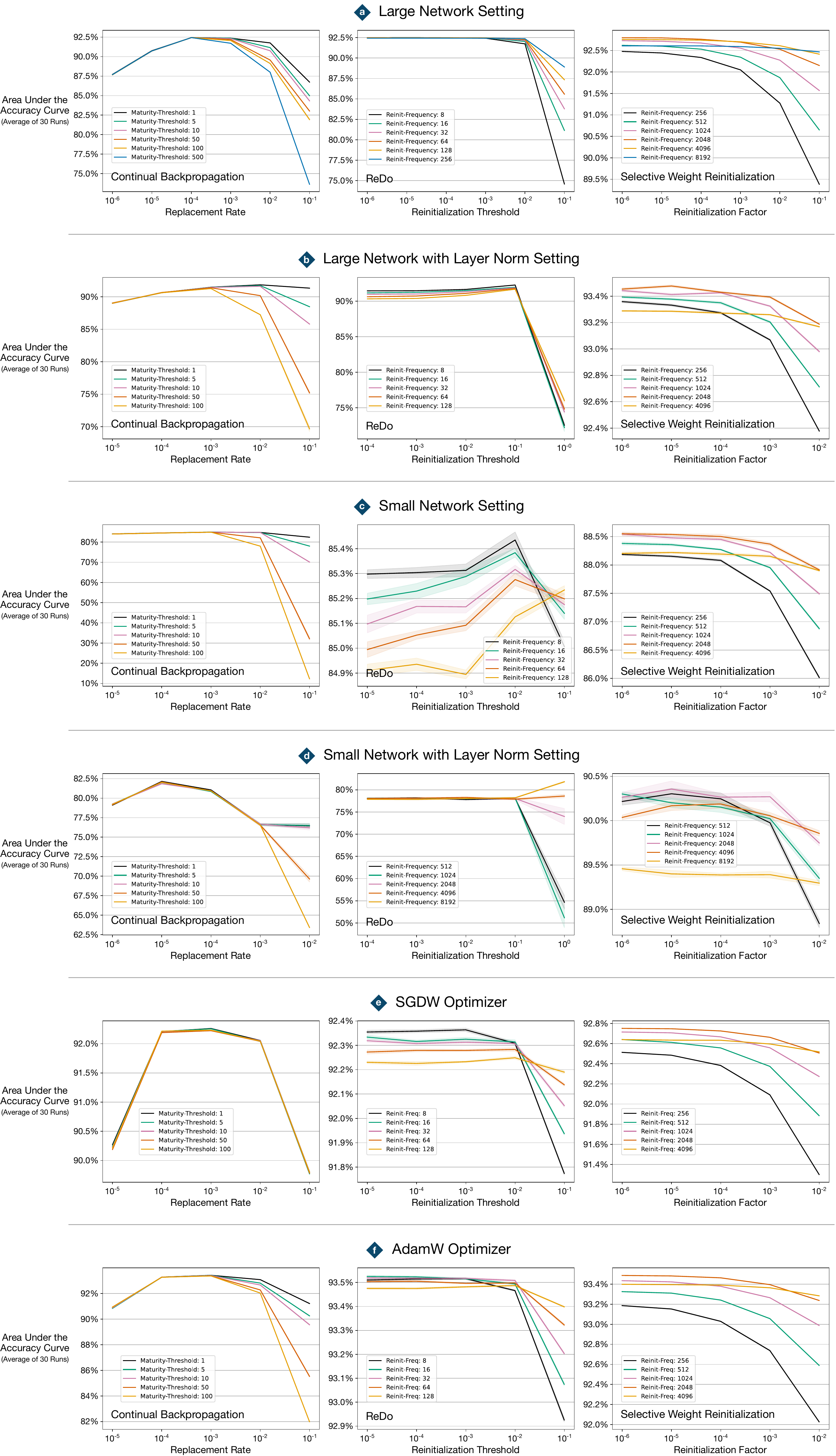}
    \caption{
    Sensitivity curve for selective reinitialization algorithms in all the settings studied in Section \ref{sec:permuted_mnist} and Appendix \ref{app:mnist_extended}.
    Each line is the average of 10 runs.
    Shaded regions correspond to the standard error.
    }
    \label{fig:sensitivity_all_settings}
\end{figure}

In Figure \ref{fig:sensitivity_all_settings}, we show the sensitivity curves for continual backpropagation, ReDo, and selective weight reinitialization with threshold pruning, gradient utility, and resample reinitialization. 
Note that in settings with layer normalization (Figures \ref{fig:sensitivity_all_settings}b and \ref{fig:sensitivity_all_settings}d) even the worst performing hyper-parameter values of selective weight reinitialization performed higher than continual backpropagation and ReDo.

\newpage

\section{Correlates of loss of plasticity in permuted MNIST}    
\label{app:correlates_mnist}                                    

We present additional measurements that could explain the performance of the learning systems presented in the main text. 
Our goal is to rule out pathological scenarios that often occur alongside loss of plasticity but are not the root cause of it.
We look for four pathological scenarios.
First, we look for a large accumulation of dead units corresponding to a loss of representational capacity in the network.
We measure the percentage of units that always output zero on a random sample of 1,500 MNIST images after applying a new permutation but before training recommences.
Second, we look for significant increases in the average magnitude of the network's parameters, which could signal instability in the optimization process.
We measure the average magnitude of the weight at the end of each task.
Third, we look for shrinkage of the average magnitude of the gradients, which could signal a slowdown in learning.
We measure the average gradient magnitude online as the network learns from new observations. 
Finally, we look for a decrease in the stable rank of the last hidden layer of the network, which corresponds to a reduction in the expressivity of the network \citep{kumar2021implicit}.
We measure the stable rank simultaneously at the end of each task, along with the percentage of dead units and the average magnitude of weight.

\begin{figure}
    \centering
    \includegraphics[width=\linewidth]{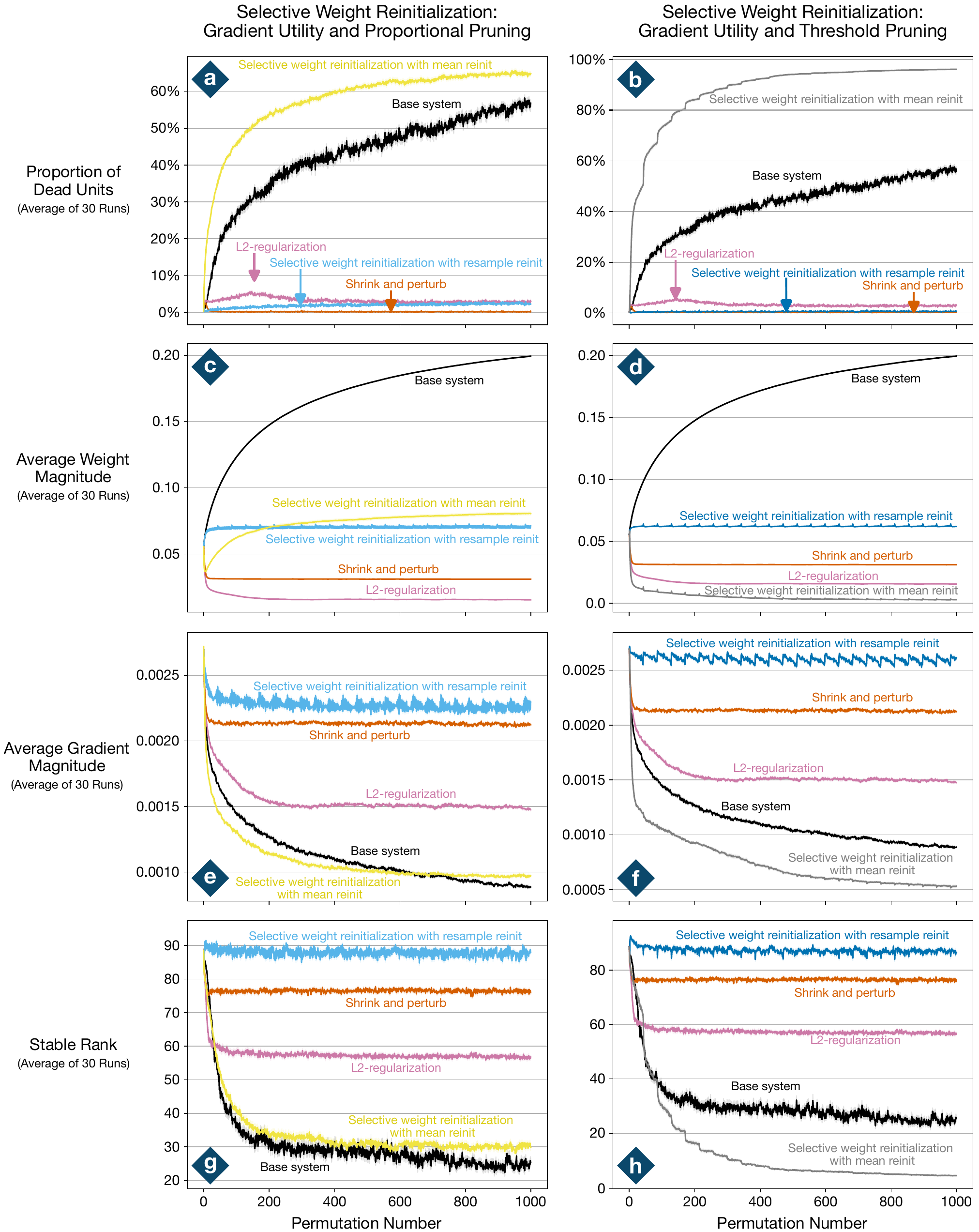}
    \vspace{-20pt}
    \caption{
    Correlates of loss of plasticity in the initial assessment of selective weight reinitialization with gradient utility.
    Each line represents the average of 30 runs, and the shaded regions correspond to one standard error. 
    }
    \label{fig:initial_assessment_correlates}
\end{figure}

\textbf{Correlates of loss of plasticity in the initial assessment.}
In Section \ref{sec:swr}, we presented comparisons between selective weight reinitialization using two pruning functions. 
While we noticed differences in performance, we did not delve deeper into the qualitative differences in the network.
Figure \ref{fig:initial_assessment_correlates} shows the four correlates of loss of plasticity.
The base system exhibited every pathological scenario we described; it had many dead units, an increasing weight magnitude, a decreasing gradient magnitude, and a decreasing stable rank. 
On the other hand, selective weight reinitialization with gradient utility and resample reinitialization (\blue and \cyan lines) avoided all of these scenarios.

\begin{figure}
    \centering
    \includegraphics[width=\linewidth]{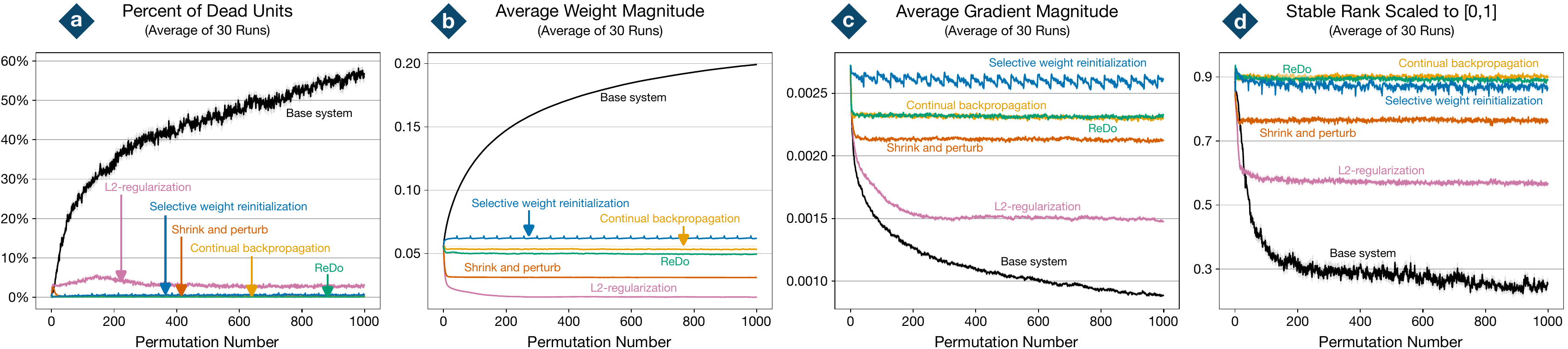}
    \vspace{-20pt}
    \caption{
    Correlates of loss of plasticity in the large network setting.
    Each line represents the average of 30 runs, and the shaded regions correspond to one standard error. 
    }
    \label{fig:large_network_correlates}
\end{figure}

\begin{figure}
    \centering
    \includegraphics[width=\linewidth]{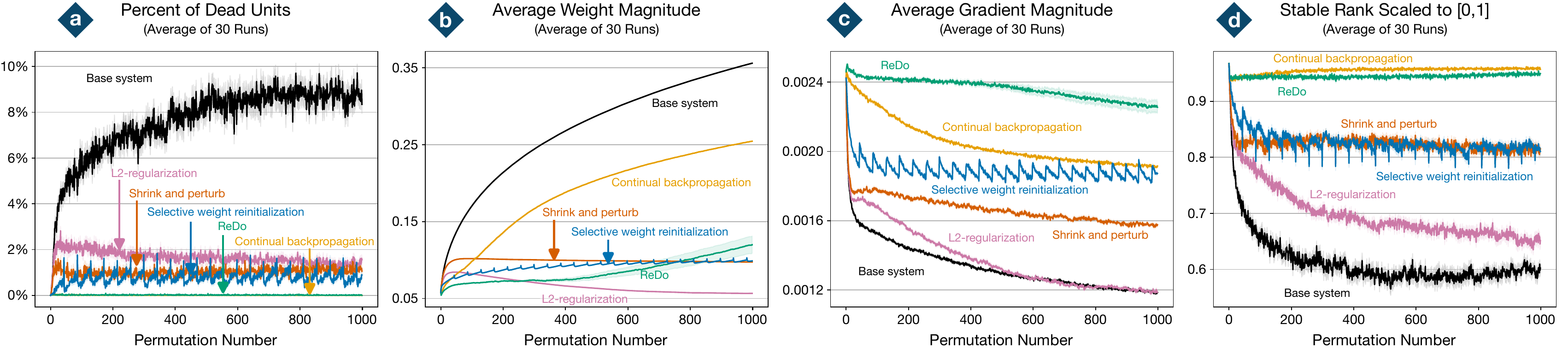}
    \vspace{-20pt}
    \caption{
    Correlates of loss of plasticity in the large network with layer norm setting.
    Each line represents the average of 30 runs, and the shaded regions correspond to one standard error. 
    }
    \label{fig:large_network_with_ln_correlates}
\end{figure}

\begin{figure}
    \centering
    \includegraphics[width=\linewidth]{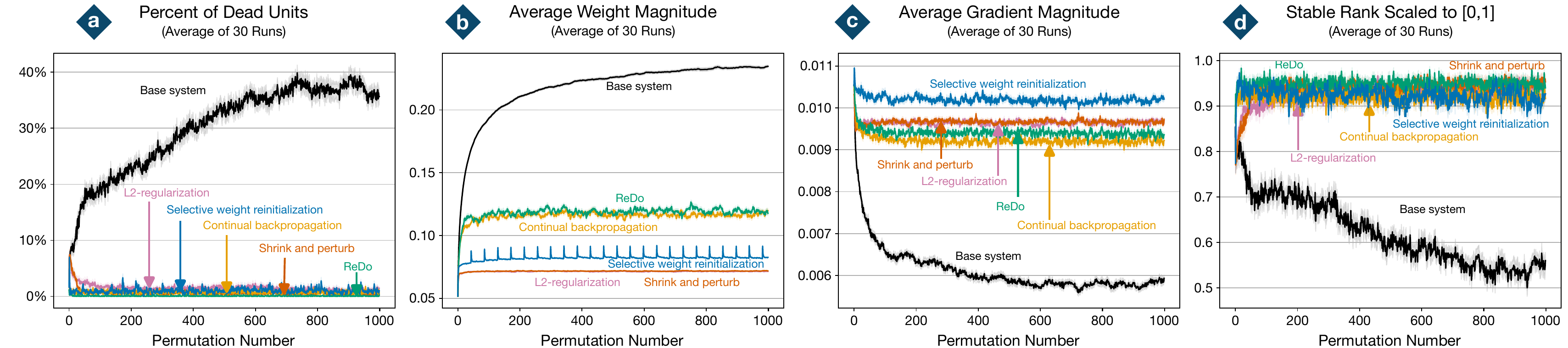}
    \vspace{-20pt}
    \caption{
    Correlates of loss of plasticity in the small network setting.
    Each line represents the average of 30 runs, and the shaded regions correspond to one standard error. 
    }
    \label{fig:small_network_correlates}
\end{figure}

\begin{figure}
    \centering
    \includegraphics[width=\linewidth]{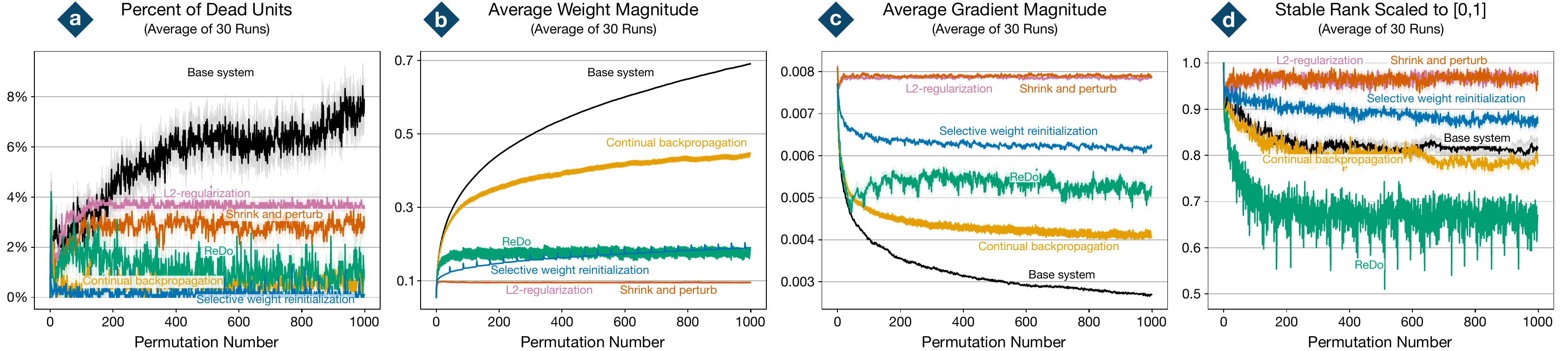}
    \vspace{-20pt}
    \caption{
    Correlates of loss of plasticity in the small network with layer norm setting.
    Each line represents the average of 30 runs, and the shaded regions correspond to one standard error. 
    }
    \label{fig:small_network_with_ln_correlates}
\end{figure}

\textbf{Comparison of the correlates of loss of plasticity of ReDo, continual backpropagation, and selective weight reinitialization.}
In Section \ref{sec:permuted_mnist}, we compared the performance of ReDo, continual backpropagation, and selective weight reinitialization with gradient utility, threshold pruning, and resample reinitialization in four different settings in the permuted MNIST problem. 
There, we discussed the differences in performance but did not look deeper into the qualitative difference in the network when using each reinitialization algorithm. 
Figures \ref{fig:large_network_correlates}, \ref{fig:large_network_with_ln_correlates}, \ref{fig:small_network_correlates}, and \ref{fig:small_network_with_ln_correlates} show the correlates of loss of plasticity for all the systems presented in Section \ref{sec:permuted_mnist}. 

All the reinitialization algorithms completely prevented the problem in terms of the number of dead units. 
In settings without layer normalization (Figures \ref{fig:large_network_correlates} and \ref{fig:small_network_correlates}), all the reinitialization algorithms maintained stable measurements in the four correlates.
However, when using layer norm (Figures \ref{fig:large_network_with_ln_correlates} and \ref{fig:small_network_with_ln_correlates}), we found that deterioration in performance was correlated with deterioration in one or more of the correlates.
For example, when using layer norm, continual backpropagation saw an increase in weight magnitude in both the large and small networks, which was also accompanied by a decrease in the magnitude of the gradient. 
The increase in weight magnitude could explain the decrease in performance due to a reduction in the effective learning rate induced by normalization layers \citep{lyle2024normalization}. 
With ReDo in the small network with layer norm setting (\green lines in Figure \ref{fig:small_network_with_ln_correlates}), we observed a decrease in stable rank and a decrease in gradient magnitude, which could explain the decline in performance observed in Section \ref{sec:permuted_mnist}.
The same effect could also explain the decrease in performance observed in the base system with regularization in the large network with layer norm setting (\pink lines in Figure \ref{fig:large_network_with_ln_correlates})

\newpage

\section{Hyper-parameter tuning and architecture in vision transformer experiment}%
\label{app:vit_tuning}                                                            %

\textbf{Hyper-parameter tuning for the vision transformer.}
We used a modified version of the vision transformer implementation in TorchVision \citep{torchvision2016}.
The architecture was modified to allow the use of ReDo and continual backpropagation.
The architecture consists of an initial convolutional layer that breaks images into patches of dimensions (4,4). 
Then, it passes through eight encoder blocks, each consisting of a layer norm module, a self-attention layer, a dropout module, a second layer norm module, and a multi-layer perceptron (MLP) block. 
The MLP block consists of a fully-connected layer, a GeLU activation, a dropout module, a second fully-connected layer, and a second dropout module. 
The ReDo and continual backpropagation layers are located after the GeLU activation, between the two fully-connected layers in the MLP block. 
We used 12 heads in each self-attention layer with a hidden dimension of 384. 
The MLP blocks used 1,536 hidden units each. 
In total, the network has 14,279,140 parameters. 

We tested different values for the highest learning rate in the schedule, $\alpha$, for the L2-regularization factor, $\lambda$, and for the dropout probability, $p$.
For the results presented in the main text, we did not scale the L2-regularization factor by the learning rate to maintain a constant regularization strength, even as the learning rate decreased to zero due to the scheduler.
When we switched to reparameterized layer norm, we kept the learning rate and dropout probability the same, but tuned the L2-regularization factor again.
The hyper-parameter values were selected to maximize test accuracy in the CIFAR-100 dataset, including all the classes over 100 epochs of training.
The network was trained using stochastic gradient descent with a momentum of 0.9 and a mini-batch size of 90.
The hyper-parameter values that we tested are given in Table \ref{tb:vit_hp}, where underlined values correspond to the ones that resulted in the highest performance. 

\begin{table}[b!]
\caption{
Hyper-parameter values for Vision Transformer on the full CIFAR-100 dataset.
}
\label{tb:vit_hp}
\begin{center}
\begin{tabular}{l c}
\multirow{3}{10em}{Regular LN \\ (base system)} 
& $p \in \{ 0.0, \underline{0.1}, 0.2 \}$ \\[1mm]
& $\alpha \in \{ 10^{-1}, \underline{10^{-2}}, 10^{-3} \}$ \\[1mm]
& $\lambda \in \{ 10^{-5}, 5 \cdot 10^{-6}, 2 \cdot 10^{-6}, 10^{-6}, 5 \cdot 10^{-7}, 10^{-7} \}$ \\[1mm]
\hdashline \\[-1.7mm]
\multirow{2}{10em}{Reparameterized LN}
& $p$ and $\alpha$ same as base system\\[1mm]
& $\lambda \in \{ 10^{-5}, 7\cdot 10^{-6}, \underline{6 \cdot 10^{-6}}, 5 \cdot 10^{-6}, 2 \cdot 10^{-6}, 10^{-6} \}$ \\[1mm]
\hdashline \\[-1.7mm]
\multirow{2}{10em}{Regular LN with no regularization on LN} 
& $p$ and $\alpha$ same as base system\\[1mm]
& $\lambda \in \{ 10^{-4}, 10^{-5}, 7 \cdot 10^{-6}, \underline{6 \cdot 10^{-6}}, 5\cdot 10^{-6}, 10^{-6} \}$ \\[1mm]
\hdashline \\[-1.7mm]
\multirow{2}{10em}{Regular LN and scaled $\lambda$}
& $p$ and $\alpha$ same as base system\\[1mm]
& $\lambda \in \{ 10^{-3}, 7 \cdot 10^{-4}, \underline{6 \cdot 10^{-4}}, 5 \cdot 10^{-4}, 10^{-4}, 10^{-5} \}$ \\[1mm]
\hdashline \\[-1.7mm]
\multirow{2}{10em}{Reparameterized LN and scaled $\lambda$}
& $p$ and $\alpha$ same as base system\\[1mm]
& $\lambda \in \{ 10^{-2}, 5 \cdot 10^{-3}, 3 \cdot 10^{-3}, \underline{2 \cdot 10^{-3}}, 10^{-3}, 10^{-4} \}$ \\[1mm]
\hdashline \\[-1.7mm]
\multirow{3}{10em}{Regular LN with no regularization on LN and scaled $\lambda$} 
& $p$ and $\alpha$ same as base system\\[1mm]
& $\lambda \in \{ 10^{-2}, 3\cdot 10^{-3}, \underline{2\cdot 10^{-3}}, 10^{-3}, 10^{-4}, 10^{-5} \}$ \\[1mm]
& \\[1mm]
\hdashline \\[-1.7mm]
\end{tabular}
\end{center}
\end{table}

In addition to the two base systems described in the previous paragraph, we also experimented with different settings for applying regularization to the network's parameters. 
As described in Section \ref{sec:lop_in_vit}, the motivation for using reparameterized layer normalization was to prevent the scaling term from shrinking and thereby stopping the flow of gradients.
This shrinkage could be due to the regularization penalty, which was applied to every parameter of the network.
Hence, another possible solution to prevent this shrinkage is to simply not penalize the learnable parameters of layer normalization.
This is a common way to use regularization in deep learning.
In practice, it is common only to regularize weight matrices and omit regularization on vectors of parameters, such as bias terms and the parameters of normalization layers \citep{goodfellow2016deep}.
Additionally, we also tested scaling the regularization factor by the learning rate, as it is standard when using learning rate scheduling.
To be specific, SGD with L2-regularization performs the update:
\begin{equation*}
    \boldsymbol{\theta}_{t+1} = (1 - \lambda \cdot \alpha_t) \boldsymbol{\theta}_t - \alpha_t\nabla_{\boldsymbol{\theta}_t} \hat{J}(\boldsymbol{\theta}_{t}),
\end{equation*}
where $\boldsymbol{\theta}_t$ are the current network parameters, $\lambda$ is the regularization factor, $\alpha_t$ is the current learning rate according to the learning rate scheduler, and $\nabla_{\boldsymbol{\theta}_t} \hat{J}(\boldsymbol{\theta}_{t})$ is the gradient of the loss with respect to $\boldsymbol{\theta}_t$.
By scaling the regularization factor, we mean multiplying $\lambda$ by $\alpha_t$ in the equation above.
When using a learning rate schedule, $\alpha_t$ decreases towards zero, which drives the regularization strength, $\lambda \cdot \alpha_t$, towards zero.
In the main text, we did not scale by the learning rate to prevent overfitting, meaning that the regularization strength was always $\lambda$. 

\begin{figure}[t!]
    \centering
    \includegraphics[width=\linewidth]{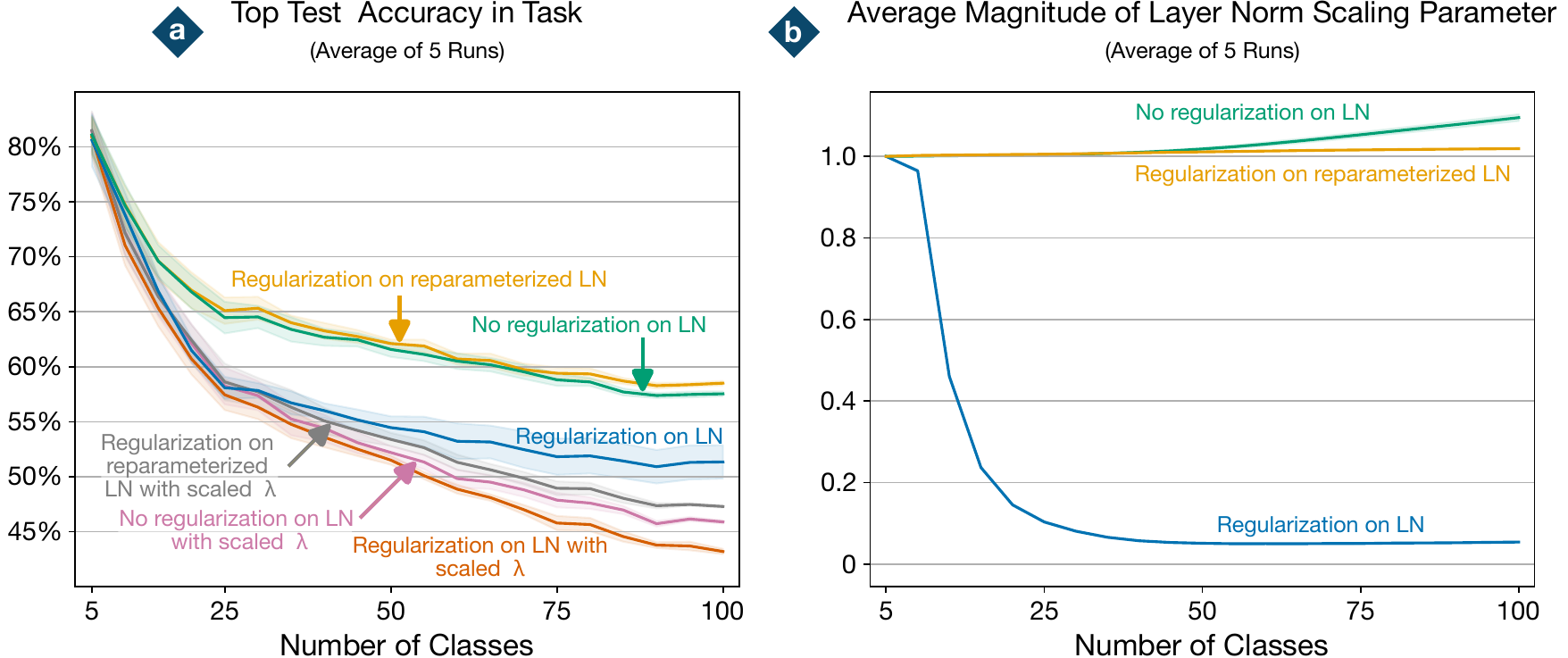}
    \vspace{-10pt}
    \caption{
    (\textbf{a}) Top test accuracy in the task for each of the learning systems with different ways of using L2-regularization.
    (\textbf{b}) Average magnitude of the scaling factor of the layer norm parameter. 
    Reparameterized layer norm without rescaling the regularization factor had the highest performance; we built upon this system in the main text. 
    Each line represents the average of 5 runs, and the shaded regions correspond to one standard error. 
    }
    \vspace{-20pt}
    \label{fig:vit_base_systems}
\end{figure}

In total, we tested three different systems with and without scaling the regularization factor by the learning rate: the network with regularization on layer norm parameters, the network with regularization on the parameters of reparameterized layer norm, and the network with no regularization on the layer norm parameters; regardless of the system, we applied regularization to all the weight matrices. 
Note that omitting regularization on the parameters of reparameterized layer norm is equivalent to omitting regularization on the parameters of standard layer norm.
For each of these systems, we tuned the regularization factor; the specific values are listed in Table \ref{tb:vit_hp}.
Note that the regularization factors, $\lambda$, for the systems with scaled $\lambda$ are orders of magnitude larger because they are eventually scaled by $\alpha_t$. 
In Figure \ref{fig:vit_base_systems}, we show the average performance of five runs for each of these systems when trained incrementally.

The systems with scaled $\lambda$ had lower performance than the systems without scaled $\lambda$, which justifies our decision of using systems without scaled $\lambda$ in the main text. 
Using reparameterized layer norm resulted in higher performance than using standard layer norm.
Omitting regularization in the layer norm parameters also resulted in improved performance, and it also prevented the scaling factor of layer norm from shrinking (\green \ line in Figure \ref{fig:vit_base_systems}b).
Note that for reparameterized layer norm in Figure \ref{fig:vit_base_systems}b, we added one to the scaling factor to account for the reparameterization. 
The standard approach in deep learning is to omit regularization on the layer norm parameters and to scale $\lambda$ by the learning rate, which corresponds to the \pink \ line in Figure \ref{fig:vit_base_systems}a. 
Yet, we achieved more than a 10\% improvement in performance from using reparameterized layer norm and not scaling $\lambda$ by the learning rate (\orange \ line in Figure \ref{fig:vit_base_systems}a). 
Nevertheless, the learning system still lost plasticity as the number of classes increased, as shown in Section \ref{sec:lop_in_vit} in the main text. 

\textbf{Effect of class order on test accuracy.}
We took several measures to ensure that overfitting was not the phenomenon responsible for the decrease in performance we observed in our experiments. 
However, one factor we did not fully account for is that some classes receive significantly more training than others.
Within a run, the first five classes are trained for a total of 2,000 epochs, while the last five classes are trained for only 100 epochs.
Thus, the network may overfit to the training data of the first initial classes, causing it to lose performance due to overfitting instead of loss of plasticity.

To check for any correlation, we ranked the test accuracy in the final task of each class from highest to lowest, i.e., from 0 to 99. 
Then, we measured the Spearman rank correlation between the ranks and the order in which each class was introduced. 
If the network is not overfitting, we should see no correlation between the order in which the classes were introduced and the rank of the accuracies in the last task.
Table \ref{tb:correlation} shows the correlation and a p-value corresponding to a two-sided permutation test. 
For systems with scaled regularization factor, there was a positive correlation, implying that classes introduced late during the experiment had higher test accuracy in the final task.
This result suggests that the networks with scaled regularization factors were overfitting to the initial classes.
On the other hand, systems without scaled regularization factor did not show any correlation. 
However, note that due to correction for multiple comparisons, we can only make these conclusions at a 10\% significance level, i.e., the rejection threshold for the p-value is $0.1 / 6 \approx 0.0167$. 

\begin{table}[t!]
\caption{
Correlation analysis between the ranks of the accuracy in last task and when classes were introduced.
}
\label{tb:correlation}
\begin{center}
\begin{tabular}{l c c}
\hline \\[-1.7mm]
\bf Learning System & \bf Spearman Correlation & \bf P-Value
\\[1mm] \hline \\[-1.7mm]
\textcolor{myorange}{Regularization on reparameterized LN}    &   0.0428  &   0.3396
\\[1mm]
\hdashline \\[-1.7mm]
\textcolor{mygreen}{No regularization on LN}                 &   0.0679  &   0.1394
\\[1mm]
\hdashline \\[-1.7mm]
\textcolor{myblue}{Regularization on LN}                    &   0.0636  &   0.1500
\\[1mm]
\hdashline \\[-1.7mm]
\textcolor{mygray}{Regularization on reparameterized LN with scaled} $\lambda$  &   0.1163  &   0.0120
\\[1mm]
\hdashline \\[-1.7mm]
\textcolor{mypink}{No regularization on LN with scaled $\lambda$}               &   0.1080  &   0.0154
\\[1mm]
\hdashline \\[-1.7mm]
\textcolor{myvermilion}{Regularization on LN with scaled $\lambda$}                  &   0.1269  &   0.0032
\\[1mm]
\hline \\[-1.7mm]
\end{tabular}
\end{center}
\end{table}

\textbf{Hyper-parameter tuning for reinitialization algorithms and shrink and perturb.}
For the three reinitialization algorithms and shrink and perturb, we followed the same procedure for tuning hyper-parameters. 
We used the sum of the highest test accuracy per task, equivalent to the area under the curve in Figure \ref{fig:vit_experiment}, as a measure of performance to guide our search. 
First, we conducted an initial random search to find a suitable hyper-parameter range. 
Second, we ran each parameter combination for five random seeds.
Finally, we selected the hyper-parameter combination that resulted in the highest average performance and ran it for an additional 15 random seeds.
We present those results in the main text in Section \ref{sec:lop_in_vit}. 

For continual backpropagation, we tested values for the replacement rate in $\{10^{-6}, 5 \cdot 10^{-7}, 10^{-7}, 5 \cdot 10^{-8}, 10^{-8}\}$ and maturity threshold in $\{10^3, 10^4\}$ for a total of ten different hyper-parameter settings.
In the main text, we present the results using a replacement rate of $10^{-7}$ and a maturity threshold of $10^4$. 
We used the contribution utility computed on the current mini-batch of data.

For ReDo, we tested values of the reinitialization frequency in $\{ 2^6, 2^7, 2^8 \}$ and the reinitialization threshold in $\{5 \cdot 10^{-3}, 10^{-3}, 5 \cdot 10^{-4} \}$ for a total of nine different hyper-parameter settings.
The parameters used in the main text were a reinitialization frequency of $2^7$ and a reinitialization threshold of $5 \cdot 10^{-4}$. 

For shrink and perturb, we tested values of $\sigma^2 \in \{ 10^{-7}, 5 \cdot 10^{-8}, 10^{-8}, 5 \cdot 10^{-9}, 10^{-9} \}$.
The value used in the main text is $10^{-8}$, which resulted in the highest performance.

For selective weight reinitialization, we used gradient utility, threshold pruning, and resample reinitialization. 
We tested values for the reinitialization frequency in $\{2^6, 2^7, 2^8\}$ and values for the reinitialization factor in $\{0^{-3}, 5 \cdot 10^{-4}, 10^{-4}\}$ for a total of nine different hyper-parameter settings.
The main text shows the results of selective weight reinitialization with reinitialization frequency and factor of $2^7$ and $5 \cdot 10^{-4}$, respectively.

\begin{figure}[t]
    \centering
    \includegraphics[width=\linewidth]{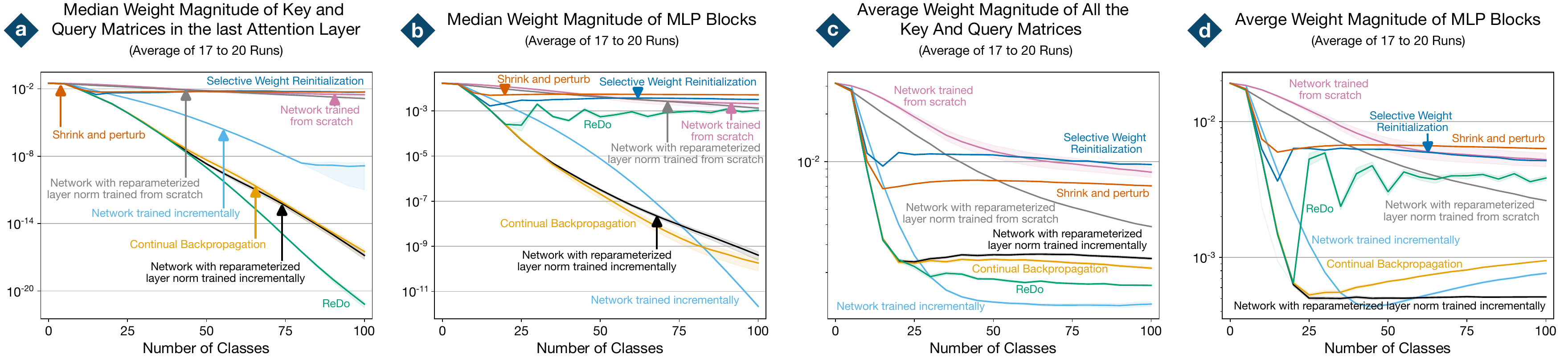}
    \vspace{-20pt}
    \caption{
    Summaries of the weight magnitude of several different layers in the vision transformer architecture: (\textbf{a}) the median weight magnitude of the key and query matrices in the last attention layer, (\textbf{b}) the median weight magnitude of the MLP blocks, (\textbf{c}) the average weight magnitude of all the key and query matrices in the network, and (\textbf{d}) the average weight magnitude of the weights in the MLP blocks. 
    We omit runs that diverged, so the measurements for ReDo are the average of 17 runs, for continual backpropagation, the average of 18 runs, and for the network with reparameterized layer norm trained incrementally, the average of 19 runs. 
    All the other lines are the average of 20 runs.
    The shaded region corresponds to the standard error.
    Note that the y-axis of each plot is on a logarithmic scale. 
    }
    \label{fig:vit_weights}
\end{figure}

\section{Learning instability in vision transformers trained incrementally}
\label{app:vit_instability}

Apart from the shrinkage of the layer norm scaling parameters, we found that several other weights in the network were reaching extremely low values. 
In extreme cases, this resulted in divergence due to underflow in one run of the network with reparameterized layer norm (black line in Figure \ref{fig:vit_experiment}b), three runs of ReDo (\green \ line in Figure \ref{fig:vit_experiment}c), and two runs of continual backpropagation (\orange \ line in Figure \ref{fig:vit_experiment}c). 
The layers that suffered the most shrinkage were the key and query matrices of the attention layers, which are located deep within the network.
For example, in Figure \ref{fig:vit_weights}a, we show the median weight magnitude of the key and query matrices in the last attention layer.
The weight magnitude in this layer decreased exponentially for ReDo (\green), continual backpropagation (\orange), the network with reparameterized layer norm trained incrementally (black), and the network trained incrementally (\cyan).
In contrast, the weights of the networks trained using selective weight reinitialization (\blue) and shrink and perturb (\vermilion) stayed relatively stable. 

In the case of continual backpropagation, it could be possible to prevent the problem if we could apply it to the attention layers.
However, we noticed that the same exponential decrease was also happening in the weights of the MLP blocks (\orange \ line in Figure \ref{fig:vit_weights}b), which do use continual backpropagation. 
ReDo seems to prevent the exponential decrease in MLP blocks, as well as selective weight reinitialization and shrink and perturb. 

Finally, these two effects are not immediately apparent across the entire network.
For example, when measuring the average weight magnitude in all the key and query matrices in the network (Figure \ref{fig:vit_weights}c), there is no evidence of such an exponential decrease.
Similarly, the average weight magnitude of the MLP blocks does not show any exponential decrease (Figure \ref{fig:vit_weights}d). 

These numerical instabilities arose from using 32-bit floats and are not inherent to the reinitialization algorithms or the network architecture. 
A simple way to avoid this problem is to use higher precision, such as 64-bit floats, which would double the memory cost of the network.
Alternatively, since weights of such small magnitude have a negligible impact on the network's output, they could be truncated to zero or clipped to a larger, yet still small, value that is safe from underflow.

\section{Extended Results with Vision Transformers}    
\label{app:vit_extended}                               

\begin{figure}[hb]
    \centering
    \includegraphics[width=\linewidth]{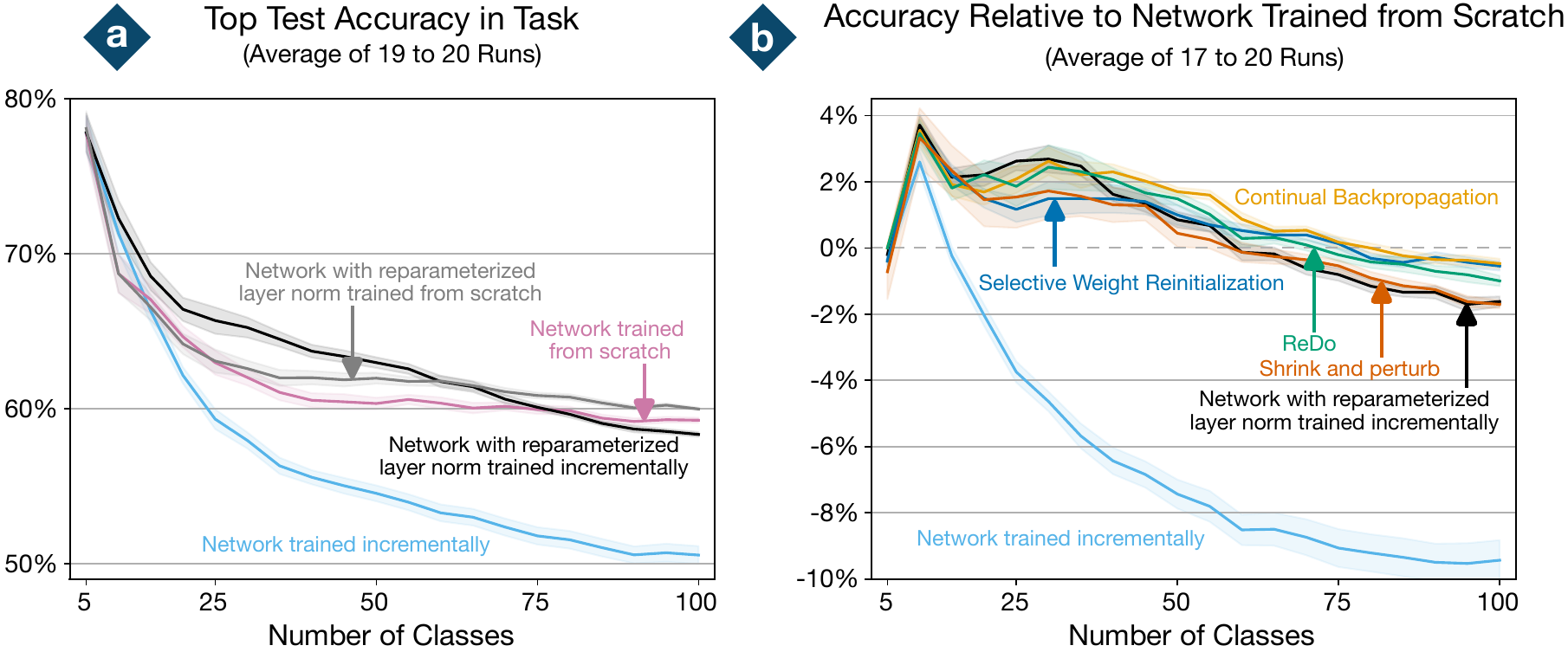}
    \vspace{-20pt}
    \caption{
    We omit runs that diverged, so the measurements for ReDo are the average of 17 runs, for continual backpropagation, the average of 18 runs, and for the network with reparameterized layer norm trained incrementally, the average of 19 runs. 
    All the other lines are the average of 20 runs.
    The shaded region corresponds to the standard error.
    }
    \label{fig:vit_filtered}
\end{figure}

\begin{figure}
    \centering
    \includegraphics[width=\linewidth]{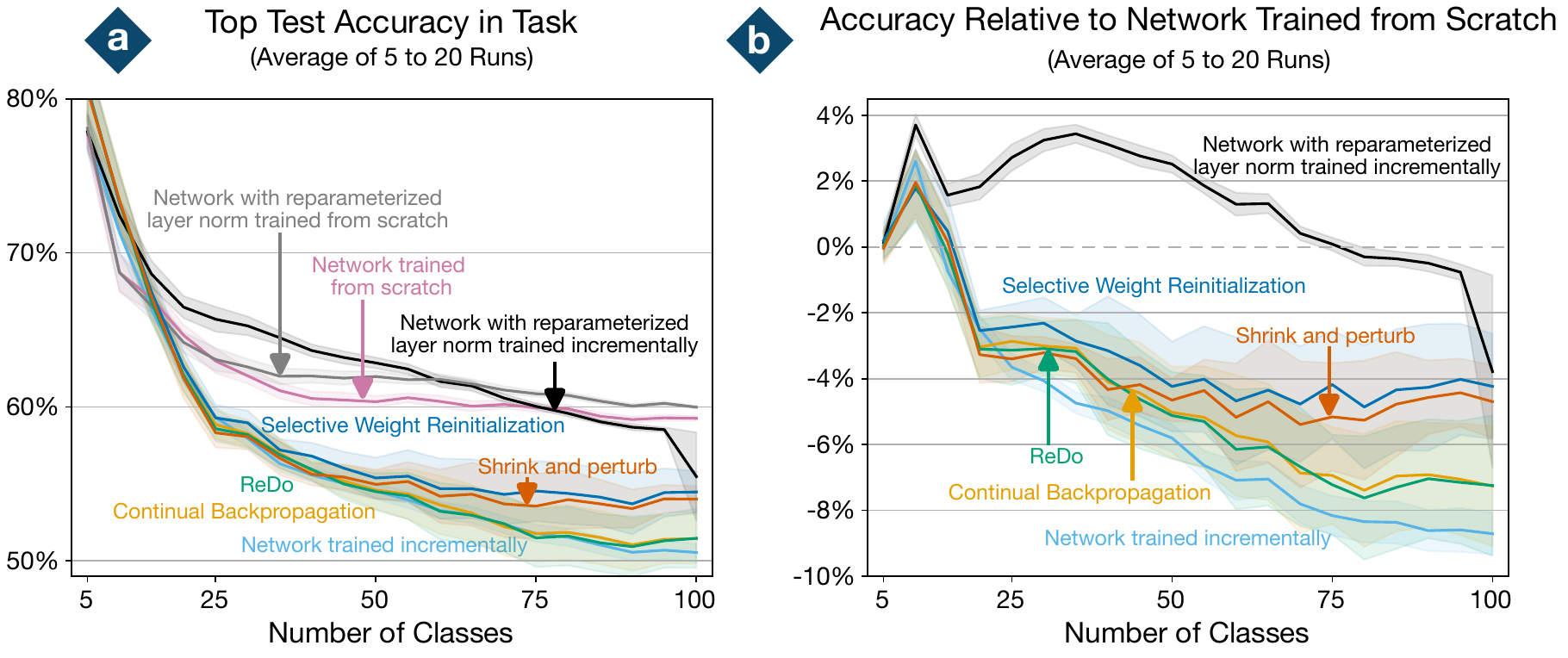}
    \vspace{-20pt}
    \caption{
    Results of reinitialization systems with standard layer norm.
    The baseline used in the difference of accuracy plot is the network trained from scratch with standard layer norm.
    The lines for ReDo, continual backpropagation, and selective weight reinitialization are the average of five runs.
    The other lines correspond to the average of 20 runs.
    Shaded regions represent the standard error. 
    }
    \label{fig:vit_no_reparameterized_ln}
\end{figure}

For completeness, we present the same results as in Section \ref{sec:lop_in_vit}, omitting diverging runs in Figure \ref{fig:vit_filtered}. 
We also include results without reparameterized layer norm in Figure \ref{fig:vit_no_reparameterized_ln}.

\end{document}